\documentclass{article}

\usepackage{microtype}
\usepackage{graphicx}
\usepackage{subcaption}
\usepackage{booktabs} 
\usepackage{wrapfig}

\usepackage{enumitem}
\setitemize{noitemsep,topsep=0pt,parsep=0pt,partopsep=0pt}

\usepackage{multirow}
\usepackage[dvipsnames]{xcolor}

\usepackage{amsmath,amsfonts,bm}

\def\eqref#1{equation~\ref{#1}}

\def\1{\bm{1}}

\DeclareMathAlphabet{\mathsfit}{\encodingdefault}{\sfdefault}{m}{sl}
\SetMathAlphabet{\mathsfit}{bold}{\encodingdefault}{\sfdefault}{bx}{n}

\usepackage{hyperref}

\usepackage[preprint]{icml2026}

\usepackage{amsmath}
\usepackage{amssymb}
\usepackage{mathtools}
\usepackage{amsthm}

\usepackage[capitalize,noabbrev]{cleveref}

\theoremstyle{plain}

\theoremstyle{definition}

\theoremstyle{remark}

\usepackage[textsize=tiny]{todonotes}

\icmltitlerunning{Random Scaling of Emergent Capabilities}

\begin{document}



\twocolumn[
  \icmltitle{Random Scaling of Emergent Capabilities}



  \icmlsetsymbol{equal}{*}

  \begin{icmlauthorlist}
    \icmlauthor{Rosie Zhao}{equal,Harvard,Kempner}
    \icmlauthor{Tian Qin}{equal,Harvard}
    \icmlauthor{David Alvarez-Melis}{Harvard,Kempner}
    \icmlauthor{Sham Kakade}{Harvard,Kempner}
    \icmlauthor{Naomi Saphra }{Harvard,Kempner}
  \end{icmlauthorlist}

  \icmlaffiliation{Harvard}{Harvard University.}
  \icmlaffiliation{Kempner}{Kempner Institute for the Study of Natural and Artificial Intelligence}

  \icmlcorrespondingauthor{Rosie Zhao}{rosiezhao@g.harvard.edu}
  \icmlcorrespondingauthor{Tian Qin}{tqin@g.haravrd.edu}
  \icmlcorrespondingauthor{Naomi Saphra}{nsaphra@fas.harvard.edu}

  \icmlkeywords{Machine Learning, ICML}

  \vskip 0.3in
]



\printAffiliationsAndNotice{\icmlEqualContribution}

\begin{abstract}
Language models famously improve under a smooth scaling law, but some specific capabilities exhibit sudden breakthroughs in performance.
Advocates of ``emergence" view these capabilities as unlocked at a specific scale, but others attribute breakthroughs to superficial metric thresholding effects. We propose that breakthroughs are instead driven by continuous changes in the \textit{probability distribution} of training outcomes when performance is bimodally distributed across random seeds. we show that different random seeds can produce \textit{either} smooth \textit{or} emergent scaling trends in synthetic length generalization tasks, multiple choice question answering, and grammatical generalization. We reveal that sharp breakthroughs in metrics are produced by underlying continuous changes in their distribution across seeds. 
These distributions may become abruptly bimodal at a capacity threshold---but this threshold appears at scales well before most seeds achieve breakthrough. 
Our observations hold true even under continuous loss metrics, confirming that random variation must be considered when predicting a model's performance from its scale.
\end{abstract}

\section{Introduction}

On most benchmarks, language model (LM) performance is determined by a scaling law \citep{hestness2017deeplearningscalingpredictable,rosenfeld2019constructivepredictiongeneralizationerror,kaplan2020scaling} that varies smoothly with parameter size and overall training compute. There are, however, a number of celebrated exceptions in which performance abruptly improves on specific benchmarks \citep{srivastava2023beyond}.
These sudden breakthroughs fuel one of the most heated debates in modern AI. 

On one side, advocates of \textbf{emergence} claim that performance abruptly improves when a particular scale provides the capacity to learn specific concepts  \citep{wei2022emergent}. On the other side, skeptics argue that these sudden improvements are a \textbf{mirage} \citep{schaeffer2024emergent} driven by thresholding effects. These threshold artifacts are alleviated by more appropriate continuous metrics---though a few \textbf{breakthrough capabilities} remain stubbornly emergent. We argue that discontinuities are driven by continuous changes in the \textit{probability} of a breakthrough at each scale. In other words, the discontinuities are real---each model firmly either \textit{knows} or \textit{does not know} a given concept---but breakthroughs do not always reflect a fixed threshold at which a concept is learnable. Instead, models may learn the concept at various scales, albeit with changing probability.

\begin{figure*}[t!]
    \centering
        \vspace{-2mm}
    \includegraphics[width=0.7\textwidth]{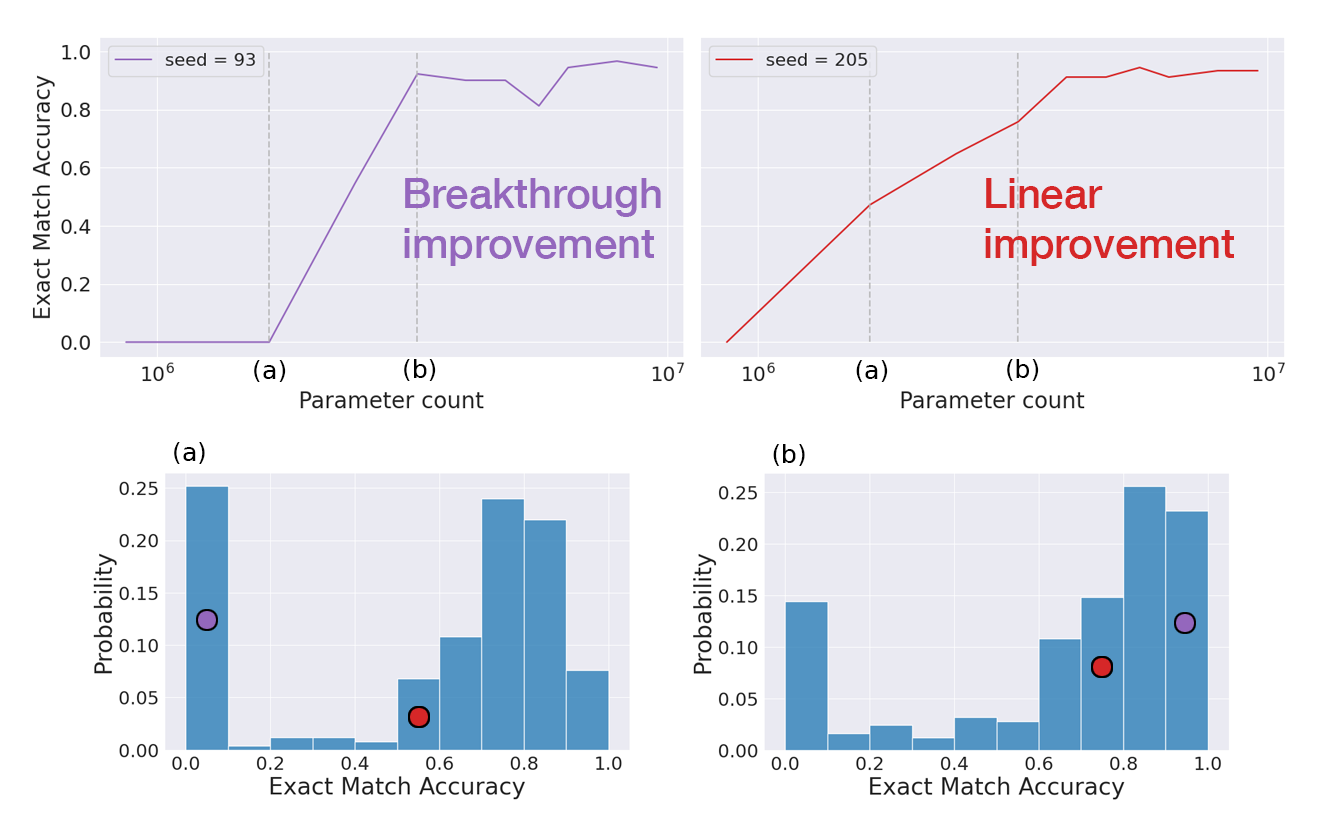}
    \vspace{-3mm}
    \caption{\textbf{Different random seeds produce different scaling trends.} Scaling trends can be emergent or linear for different seeds, even if all models train on the same data with the same hyperparameters. On the count task (see Appx. \ref{app:count}), we show trends for random seeds with the highest breakthroughness (seed 93; top left) and linearity (seed 205; top right). We mark parameter counts immediately before and after seed 93's emergence, respectively, as (a) and (b). Histograms illustrate the bimodal distribution of performance across all random seeds at scales (a) and (b), marking the positions of seeds 93 and 205. Breakthroughs occur when consecutive points represent different clusters; linear trends occur when each point is sampled from the same gradually shifting cluster.}
    \label{fig:emergence}
\end{figure*}

We posit that a breakthrough capability is  distinguished not by deterministic responses to scale, but by \textit{multimodal} random variation. In other words, independent training runs cluster in their performance metrics. This observation is undocumented because scaling laws usually plot a single training run at each scale, rarely testing multiple seeds. Although random variation may be benign when model performance is measured in-distribution~\citep{jordanvariance}, previous work suggests that out-of-distribution performance may vary widely across training runs~\citep{zhoualgorithms, zhou2024transformers,qin2024itree,juneja_linear_2023,li2025interpretationpredictbehaviorunseen}, even at larger scales~\citep{madaan2024quantifying}. 

By connecting breakthrough scaling with random variation, we challenge the narratives of both the \textit{emergence} and \textit{mirage} camps. First, our results \textit{contest the position of the mirage or ``loss-to-downstream'' camp}: that effects of scale are predictable and that continuous metrics will smooth out apparent breakthroughs. We discover clustered multimodal distributions of capabilities, confirming that models unpredictably learn critical discrete concepts---and these clusters are observable even using continuous loss metrics.  Furthermore, we \textit{complicate the rival narrative of the emergence camp}, which often treats specific model sizes as distinct in their capacity. While we confirm that algorithms require some minimum model capacity, that capacity may not be the enormous scale at which emergence is observed. Instead, \textit{discontinuous} performance jumps are sampled from a \textit{continuously} changing multimodal distribution, where the ``success'' mode ultimately dominates at larger scales. To extrapolate downstream metrics to large scales, we must predict the likelihood of successful emergence as well as the performance of a successful run.

Since training numerous seeds is prohibitively expensive at large scales, we study partially reinitialized LLMs and toy models. Whereas prior work reports summary statistics across only a few training runs, we characterize the full multimodal performance distribution. We find: 

\begin{figure*}[t!]
    \centering

    \begin{subfigure}{\textwidth}
        \centering
        \includegraphics[width=\linewidth]{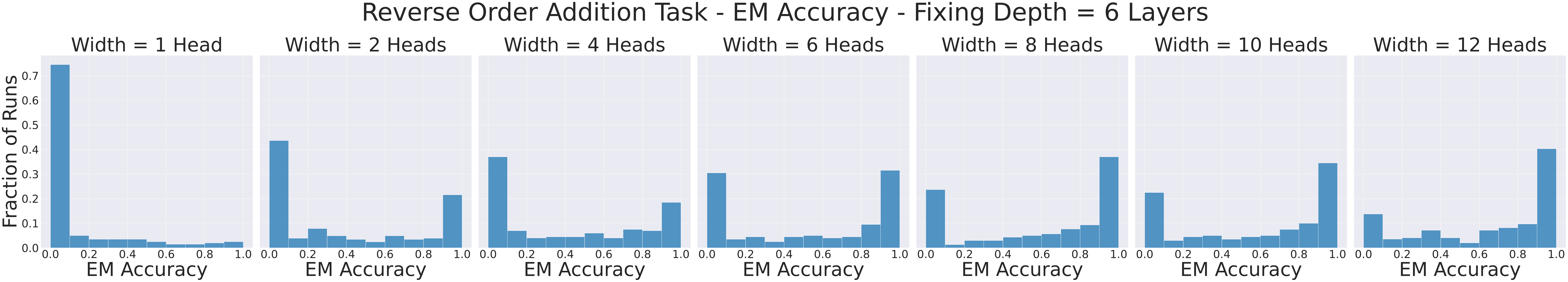}
        \caption{Fixing network depth to 6 layers.}
        \label{fig:rev_add_fix_depth_hist}
    \end{subfigure}

    \vspace{0.2cm}

    \begin{subfigure}{\textwidth}
        \centering
        \includegraphics[width=\linewidth]{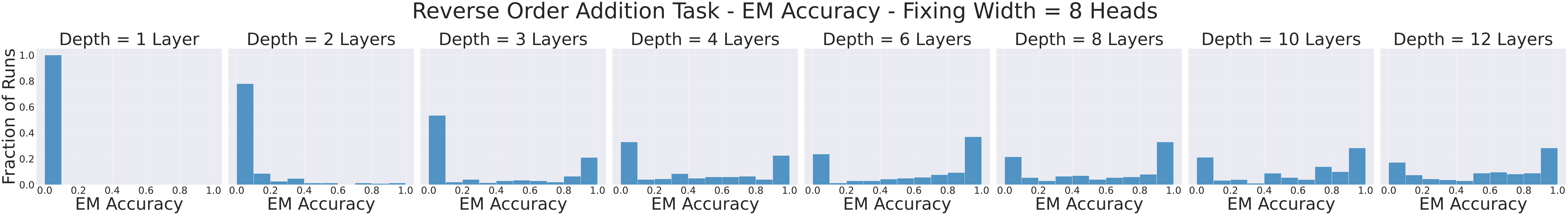}
        \caption{Fixing network width to 8 heads per layer.}
        \label{fig:rev_add_fix_width_hist}
    \end{subfigure}

    \caption{\textbf{Random variation in length generalization (addition task).}
    Histograms of exact match accuracy on length $40$ sequences when independently scaling
    \subref{fig:rev_add_fix_depth_hist} width and \subref{fig:rev_add_fix_width_hist} depth.}
    \label{fig:rev_add_histograms}
\end{figure*}

\begin{itemize}
\item \textbf{Breakthroughs result from bimodal performance distributions.} On synthetic algorithmic tasks (Sec. \ref{sec:synthetic-results}), simple language modeling tasks (Sec.~\ref{sec:grammar}) and multiple choice question answering (Sec. \ref{sec:natural_language}) , some random seeds produce linear scale trends while others are emergent. This variation is caused by the bimodal distribution of each skill across  seeds (Sec. \ref{subsec:bimodality}), a property that materializes around \textbf{breakthrough thresholds} in model size. At these scales, emergence is a stochastic property (Sec. \ref{subsec:stochastic-emergence}).

\item \textbf{Bimodal variation persists under continuous metrics.} In the emergence debate, one main position \citep{schaeffer2024emergent} is that breakthroughs are caused by measuring discontinuous metrics such as exact match accuracy rather than loss. We find continuous loss metrics can remain visibly bimodally distributed, particularly when there is a more even split of failed and successful runs. We confirm these results on both synthetic (Sec. \ref{subsec:bimodality_mirage}, Sec.~\ref{sec:qf_result}) and natural (Sec. \ref{subsec:mmlu-mirage}) tasks.
\item \textbf{When a scale curve exhibits sudden \textit{discontinuous} improvement in a skill, the \textit{probability} of learning that skill may be changing \textit{continuously}}. Treating the bimodal distribution as a mix of \textbf{failure} and \textbf{success} distributions, we illustrate that average improvements can come from changes in the probability of success \textit{or} in the mean performance of a successful run (Sec.~\ref{subsec:bimodality}). Although bimodality can appear abruptly at a minimum capacity scale (Sec.~\ref{subsec:emergence_of_bimodality}), these sudden distributional changes do not necessarily align with breakthrough scales for individual model runs.   

\item \textbf{Bimodality can produce nonmonotonic scaling trends.} We discuss a synthetic task with ranges of inverse scaling in mean accuracy (Appx.~\ref{subsec:inverse_scaling}). The average of \textit{successful} runs remains monotonic in these ranges, suggesting that U-shaped curves in overall mean are an artifact of success probability.


\end{itemize}

\section{Synthetic Length Generalization Tasks}
\label{sec:synthetic-results}



Usually, scaling trends are measured on a single model per scale or, at most, the average of a few runs. The literature suggests that emergent capabilities unlock at specific model scales~\citep{wei2022emergent}, implying that different training runs would perform similarly at each scale. Contrary to this belief, we demonstrate that performance is sampled from a stochastic distribution which changes gradually even as individual scaling curves jump abruptly.

\subsection{Tasks and Setup}
\label{sec:methodology_synthetic}

After training models on synthetic tasks, we measure their length generalization \citep{graves2016hybrid, kaiser2015neural, lake2018generalization, hupkes2020compositionality}, one of many compositional skills that can lead to conceptual breakthroughs \citep{srivastava2023beyond,löwe2024abruptspontaneousstrategyswitches,chen_sudden_2024}. Experimental details are in Appx.~\ref{app:exp_details}.

\textbf{Architecture:} In our synthetic experiments, we train decoder-only Transformer models from scratch  using rotary position embeddings (RoPE)~\citep{su2024roformer}. To observe the random performance distribution at each scale, we train our models from hundreds of seeds. We choose model sizes by separately adjusting either the width (number of 64-parameter heads per layer) or depth  hyperparameter.

\textbf{Task: } We consider two algorithmic tasks previously studied in \citet{zhoualgorithms}: counting and addition.
An analysis of the \textbf{count} task is provided in Appx. \ref{app:count}, including discussion (Appx. \ref{subsec:inverse_scaling}) of how its average accuracy responds non-monotonically to scale. 
In the main body of this paper, we will focus on the \textbf{addition} task.
\citet{zhou2024transformers} showed that Transformers can generalize 10-15 digits past training length for an addition task, if provided with index hints and allowed to generate the answer backwards. 
We use this modified reverse-order addition task.

\textbf{Dataset:} During training, we sample sequences i.i.d from the train set and invoke in-context learning by adding examples to the context, following prior work~\citep{jelassirepeat, zhou2024transformers}. The lengths of examples are sampled uniformly from 1 to the maximum training length (30 for count and 35 for addition). Length generalization is then tested at length 60 for count and 40 for addition.

\subsection{Emergence is a Sign of Bimodal Variation}
\label{subsec:bimodality}

What do the scaling curves for length generalization tasks look like when we generate them in conventional ways? Following \citet{srivastava2023beyond}, we calculate the \emph{breakthroughness} and \emph{linearity} of scaling curves where each run shares a fixed initialization and shuffle seed.\footnote{Although it is standard practice to fix the random seed when comparing LM benchmark performance across scales, the initializations produced by a single seed have no meaningful relation between different model sizes.} As defined in Appx.~\ref{app:breakthrough}, breakthroughness measures emergence, whereas linearity measures a smooth response to scale. We plot the performance across scale for the seeds with the highest breakthrough and highest linearity in Fig. \ref{fig:emergence} (with other extremes shown in Appx. Fig.~\ref{fig:breakthrough_seeds}). Because breakthroughs vary across seeds, we can easily find fixed seeds that lead to \textit{either} emergent \textit{or} smooth scaling. 

This variation is explained by the bimodality of model performance distributions when varying seed. Fig.~\ref{fig:rev_add_histograms} illustrates that, for a population of models independently trained on the addition task, length generalization ability clusters into high and low component modes at many parameter sizes. This clustering produces distinctly bimodal performance distributions, causing some model runs to appear as breakthroughs while others generalize poorly. When bimodally distributed runs cluster into distinct high- and  low-performance components, a model might exhibit linear scaling if sampled from the same cluster as the previous scale \emph{or} emergent scaling when switching from the low cluster to the high cluster. These differences ultimately lead to high variability in the timing and degree of emergence. Furthermore, these differences cause oscillating scale curves like those seen in low-linearity tasks \citep[][ref. Fig. 7c]{srivastava2023beyond}.

\begin{figure}[h]
        \centering
        \includegraphics[width=\linewidth]{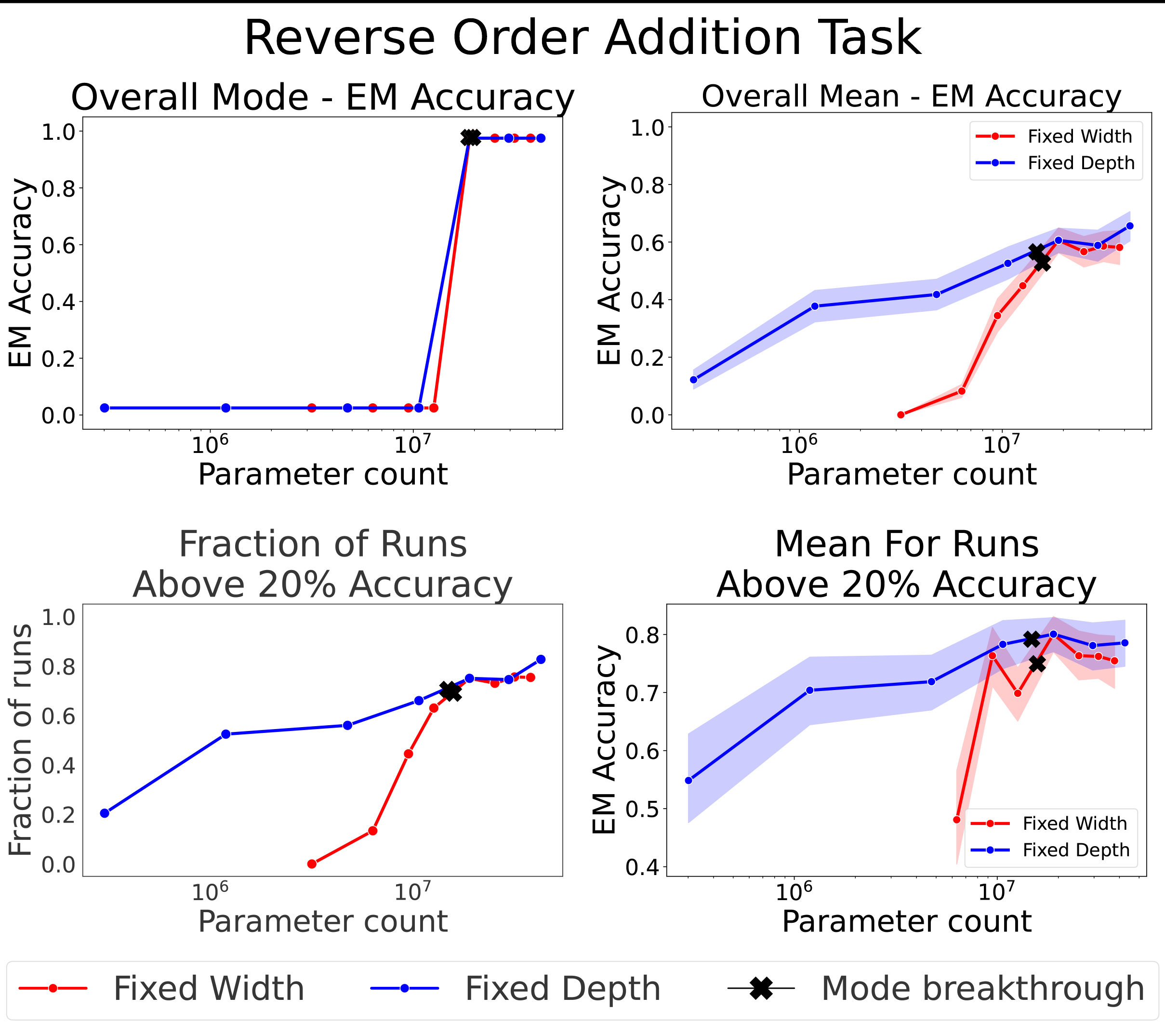}
    \caption{\label{fig:rev_add_quantiles} \textbf{Summary statistics for length generalization accuracy on the reverse-order addition task.} We track overall mode (top left) and overall mean (top right). While the mode exhibits a sharp increase in accuracy at a certain scale of width or depth, the mean evolves more continuously as a result of the bimodal nature of the random variation distribution. We also study successful runs, plotting the fraction of runs reaching above 20\% accuracy (bottom left) and the mean of such runs (bottom right). The plots for mean include 95\% confidence intervals with 1000 bootstrapped samples.}
\end{figure}

\subsection{Sudden Jumps From Gradual Distribution Shifts}
\label{subsec:stochastic-emergence}

When reporting metrics from only one seed or the mean of a few seeds, the  outcome is likely to be close to the \emph{mode} performance of the underlying model population. In Fig.~\ref{fig:rev_add_quantiles} (top left), the mode performance shows a massive spike, mirroring an  emergent benchmark's scaling trend. However, we claim that this discontinuous improvement in mode---as well as the discontinuities seen in single-seed emergent scaling curves---is an artifact of bimodality.  Underlying this \textit{discontinuous} performance jump are \textit{continuous} changes in other distributional statistics. In Fig.~\ref{fig:rev_add_quantiles} (top right), the mean exhibits a smoother trend in accuracy. Mode and mean differ because the underlying distribution is bimodal, expressing a mixture of ``successful'' and ``failing'' runs. 

Treating the distribution as a mixture of successes and failures, we can separately analyze the \textit{probability} of a successful run and the \textit{performance distribution} of successful runs. Both of these properties are changing continuously and gradually when the mode increases abruptly. If we restrict our analysis to the runs achieving at least nontrivial 20\% accuracy, we see that the probability (Fig.~\ref{fig:rev_add_quantiles} (bottom left)) and mean  (bottom right) of such ``successful'' runs both exhibit continuous improvement, with the exception of increasing from depth 2 to 3 (discussed further in Sec.~\ref{subsec:emergence_of_bimodality}). Even at the mode breakthrough, these underlying distributional properties are only changing gradually. We conclude that \textit{when tasks exhibit bimodal 
distributions across random seeds, gradual statistical improvements cause seemingly abrupt improvements with scale.}

\subsection{Is Bimodality a Mirage?}
\label{subsec:bimodality_mirage}

Metrics with hard thresholds can artificially induce breakthroughs \citep{schaeffer2024emergent}; conversely, continuous metrics turn apparent emergence into smooth curves \citep{srivastava2023beyond}. We must be particularly cautious about claiming emergence when requiring outputs to exactly match a target string, as we have so far. Are our case studies artifacts of thresholding effects? Do our bimodal distributions become unimodal under continuous metrics? 

To avoid thresholding artifacts, consider a continuous equivalent to the exact match metric: the maximum loss assigned to any individual token. Each token is individually computed in the correct output context, so the entire sequence represents a fixed set of continuous per-token loss functions. Because the maximum of a fixed set of Lipschitz-continuous functions is always Lipschitz-continuous, this error score is guaranteed to be continuous as long as per-token loss is continuous. 
For model $f$ on dataset $X$, the continuous error score based on per-token loss $L(f(x_{0 \ldots i-1}), x_i)$ is:
\begin{align}
\label{eq:continuous_error}
    \textrm{error}(f, X) = \frac{1}{|X|} \sum_{x \in X} \max_{i < |x|} L(f(x_{0 \ldots i-1}), x_i)
\end{align}
In Fig.~\ref{fig:combined_results}~\textit{(a)(b)}, we plot this continuous metric for addition models on the length generalization dataset. The distribution of this  metric across random seeds is still clustered. We therefore confirm that \textit{emergent capabilities exhibit bimodal  distributions even when using a continuous performance metric}; their bimodality is not due to thresholding alone. Further findings from these experiments are in Appx.~\ref{app:add_continuous_metrics}.

\subsection{Bimodality Emerges Abruptly}
\label{subsec:emergence_of_bimodality}

\begin{figure*}[t]
    \centering
    \begin{minipage}{0.62\textwidth}
        \centering
        \begin{subfigure}{\linewidth}
            \centering
            \includegraphics[width=\linewidth]{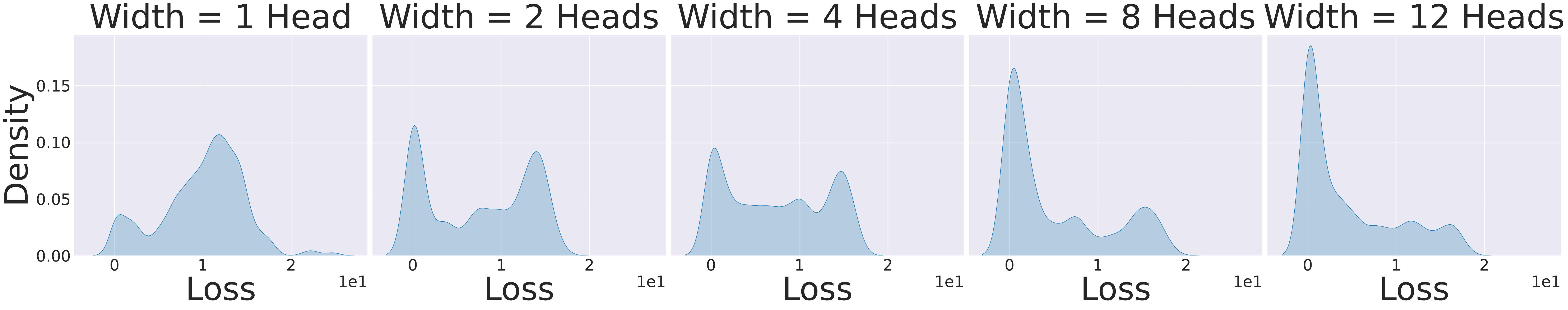}
            \caption{Fixing network depth to 6 layers while scaling up width.}
            \label{fig:rev_add_histograms_loss:depth}
        \end{subfigure}
        \vspace{2mm}
        \begin{subfigure}{\linewidth}
            \centering
            \includegraphics[width=.6\linewidth]{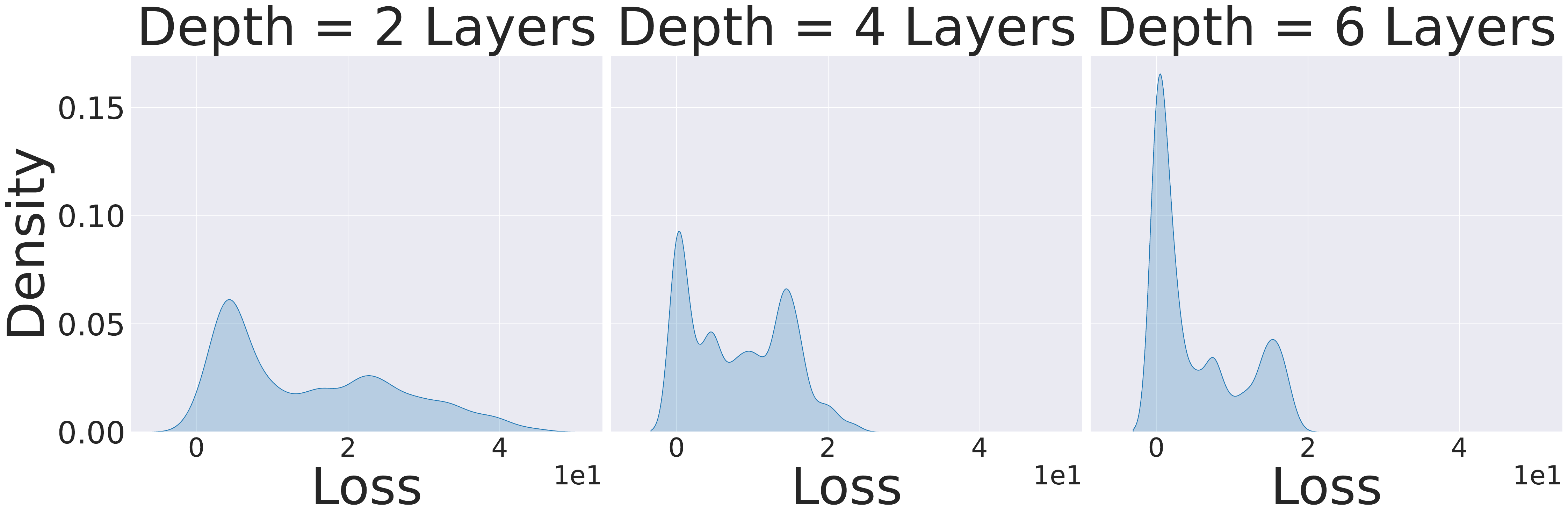}
            \caption{Fixing network width to 8 heads per layer while scaling up depth.}
            \label{fig:rev_add_histograms_loss:width}
        \end{subfigure}
    \end{minipage}
    \hfill
    \begin{minipage}{0.35\textwidth}
        \centering
        \includegraphics[width=\linewidth]{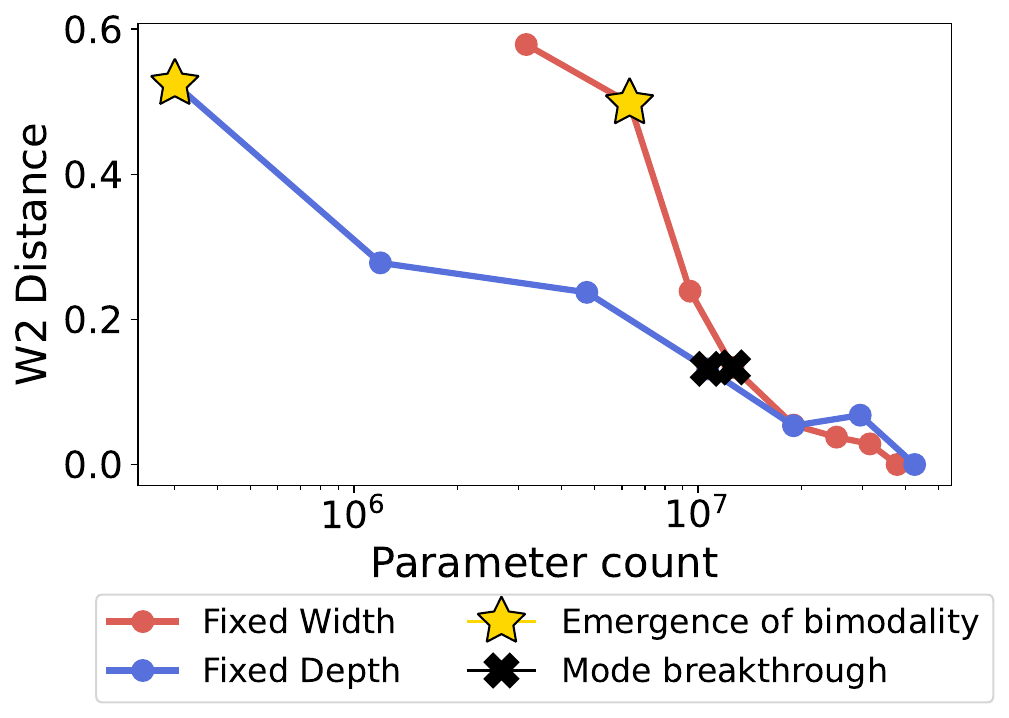}
        \caption*{\small (c) Changes in random variation.}
        \label{fig:rev_add_wasserstein}
    \end{minipage}

    \caption{\textbf{Analysis of performance distributions (addition task).} 
    \textbf{Left (a, b):} Kernel Density Estimation (KDE) of our loss-based error metric (Eq.~\ref{eq:continuous_error}). At scales where accuracy is bimodal, the distribution remains bimodal even with a continuous metric. 
    \textbf{Right (c):} Wasserstein-L2 distance of each scale's performance relative to the largest scale. We mark the ``emergence of bimodality'' and the ``mode breakthrough.'' During breakthrough, distributions shift gradually.}
    \label{fig:combined_results}
\end{figure*}

The previous section showed that bimodal performance distributions produce emergent scaling curves for \textit{individual seeds}. We next show that the \emph{distribution itself} can change suddenly at certain scales. As shown in Fig.~\ref{fig:rev_add_histograms}, the performance distribution starts as unimodal at the smallest scale, where no models can length generalize. Larger scales yield a bimodal distribution where most of the probability mass is ultimately placed on successful (i.e., length-generalizing) runs. A priori, there could be a smooth evolution between these two distributions in which probability mass from failing runs gradually shifts towards higher performance metrics, eventually splitting into a clear separate cluster. We find instead that the shift from unimodal to bimodal is abrupt and instantly polarized into low and high clusters. In Appx. Fig.~\ref{fig:count_fix_depth_1}, we confirm that models fail to generalize when fixing depth to be one layer even when scaling width (despite having perfect in-distribution accuracy), illustrating that the length-general function is suddenly learnable at sufficient depth.

To track the evolution of the performance distribution across scales, we plot the Wasserstein-L2 distance of each scale's distribution relative to that of the final, largest model scale. In Fig.~\ref{fig:combined_results}\textit{c} we see a sharp decrease in the W2 distance, marking the sudden appearance of highly successful runs when model depth reaches 3 layers or model width reaches 2 heads. 
These sudden changes identify the moment when a new capability is unlocked, as the distribution transitions abruptly from unimodal to bimodal. We posit that this transition marks the \textbf{minimum capacity} required to learn the task. We also mark the point in each trend where the mode in Fig.~\ref{fig:rev_add_quantiles} (top left) increases sharply. We argue that one can draw misleading conclusions about minimal model capacity when only studying using single runs at each scale, whereas \textit{distributional statistics correctly identify the minimum capacity that a skill requires}.

\section{Large Language Model Experiments}
\label{sec:natural_language}

After connecting emergence with bimodality in small synthetic settings, we turn to LMs. We focus on the MMLU multiple-choice question answering dataset, where performance jumps \citep{srivastava2023beyond} after the LM learns the multiple choice format \citep{hu2024auxiliary}. We will show that emergent scaling curves in LMs express underlying multimodal performance distributions.

\subsection{Data and Setup}
\label{sec:mmlu_method}

To avoid the expense of repeatedly training large LMs from scratch, we simulate independent runs by reinitializing the upper layer of pretrained LMs before continuing to train them. Our continued training dataset, a mix of C4 news data and MMLU training data, is insufficient to recover the base model's full MMLU performance. We hypothesize that for compute scales near the emergence threshold, performance distributions will be bimodal across random seeds.

During pretraining, a large diverse corpus encourages models to acquire various capabilities. While sufficiently large models may learn all capabilities, smaller models have limited capacity, requiring capabilities to compete~\citep{merrill2023tale}. This competition, influenced by initialization and data order, leads to varying outcomes across random seeds, forming performance clusters.

\textbf{Task}: We test LMs on MMLU \citep{hendrycks2021measuring}, a multiple-choice question-answering benchmark. Strong MMLU performance requires (1) natural language reasoning with domain knowledge and (2) producing answers in the required format. The latter drives emergent trends \citep{srivastava2023beyond,hu2024auxiliary}.

\textbf{Model}: We use the Qwen2.5 family of base models \citep{qwen2.5}. To introduce randomness, we reinitialize the final attention layer and LM head, then perform full-parameter continued pretraining.

\textbf{Data}: We mix the C4 news subset \citep{raffel2023exploring} with MMLU training data to ensure the multiple choice formatting circuits compete with general language modeling. We vary the MMLU proportion to control task-specific data size.\looseness=-1

\textbf{Training}: We continue pretraining Qwen2.5-0.5B and Qwen2.5-1.5B on C4-MMLU mixes, training 80 reinitializations per data mixture ratio. We train for 2 epochs (0.5B) and 5 epochs (1.5B) with learning rate 1e-5 and linear decay.

\begin{figure*}[ht]
    \centering
    \begin{subfigure}{0.4\textwidth}
        \centering
        \includegraphics[width=0.9\linewidth]{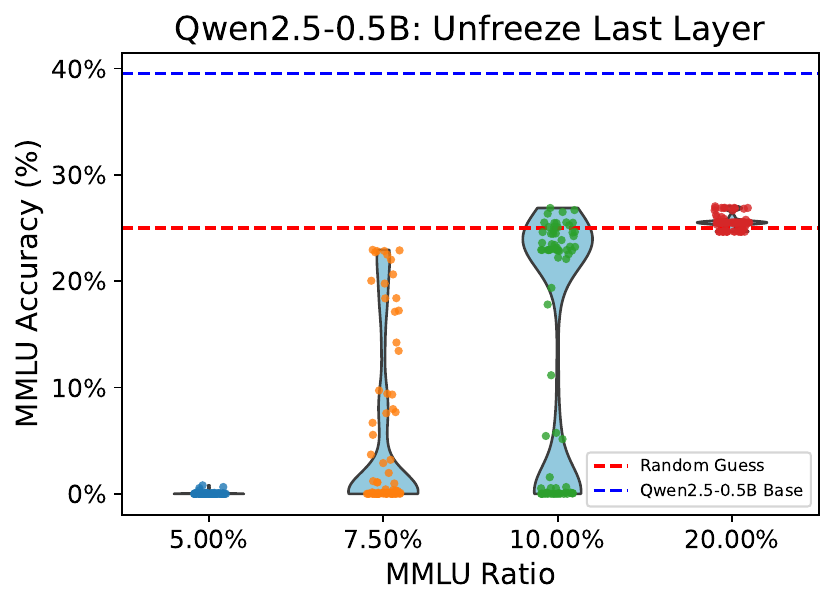}
        \caption{Qwen2.5-0.5B mix results.}
        \label{fig:qwen_emergence:data}
    \end{subfigure}
    \hspace{10px}
    \begin{subfigure}{0.4\textwidth}
        \centering
        \includegraphics[width=0.8\linewidth]{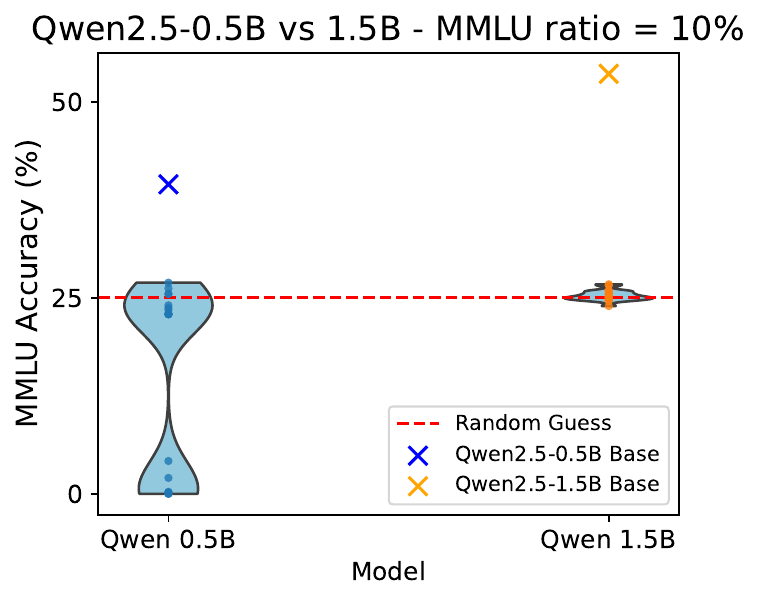}
        \caption{Qwen2.5: 10\% MMLU data.}
        \label{fig:qwen_emergence:size}
    \end{subfigure}
    \caption{
    \textbf{Performance on MMLU.} All models have their last layer reinitialized randomly and then are trained on a mix of C4 news and MMLU, with varying proportions of MMLU.
    \textbf{(a)} With insufficient MMLU training data, performance is trivial; as data increases, performance becomes bimodal across seeds. 
    \textbf{(b)} For a fixed data mix, smaller models show bimodal performance while larger models consistently relearn the task, albeit below original base levels.
    }
    \label{fig:qwen_emergence}
\end{figure*}

\subsection{Experiment Results}

\paragraph{Emergence Across Data Compositions.}

When training data contains more MMLU examples, the Qwen2.5-0.5B performance distribution improves in ways that mirror the effect of scale.
In Figure~\ref{fig:qwen_emergence} (\textit{left}), when trained on 5\% MMLU and 95\% C4 \citep{raffel2023exploring}, most models (out of 80 seeds) achieve near 0\% MMLU accuracy, failing to process the multiple-choice format. At 10\% MMLU training data, bimodality emerges: one cluster remains near 0\% accuracy (format failure), while the other reaches $\geq$25\% accuracy (random baseline). This second cluster contains models that consistently produce valid multiple choice responses; those exceeding 25\% have learned to compose format following with some recovered world knowledge. Due to the limited scale and quality of our continued pretraining data, fully recovering the base model's MMLU capability is challenging. At $>20\%$ MMLU, all 80 seeds consistently perform around or above the random baseline. Qwen2.5-1.5B shows the same bimodality (Figure~\ref{fig:qwen_mmlu_1.5b}), but the larger model recovers similar capabilities at much lower data mixture ratios.

\paragraph{Emergence Across Model Scales.}

Model size affects the clustered MMLU performance distribution similarly to the synthetic settings in Sec. \ref{sec:synthetic-results}. Using 10\% MMLU training data, we continually pretrain Qwen after reinitializing with 80 different seeds (Figure~\ref{fig:qwen_emergence}, \textit{right}). While Qwen2.5-0.5B forms two distinct performance clusters, Qwen2.5-1.5B consistently performs around or above the random baseline. Larger models reliably acquire MMLU capability on the same dataset that produces highly bimodal variation at smaller scales, consistent with scaling laws: smaller models require more training examples to match larger model performance \citep{rosenfeld2019constructivepredictiongeneralizationerror,kaplan2020scaling}.

\begin{figure*}
    \centering
    \includegraphics[width=1.0\linewidth]{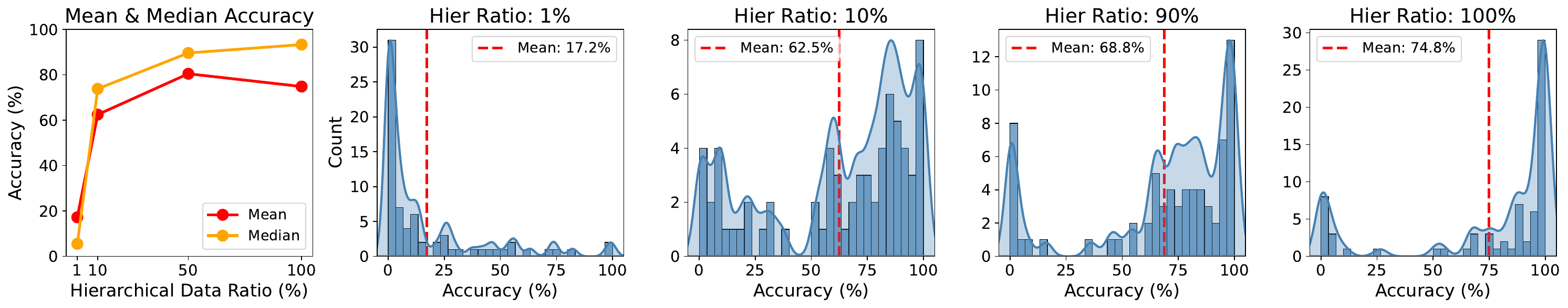}
    \caption{
    \textbf{Question formation performance across data compositions.} \textit{Left}: Mean and median OOD accuracy across 80 seeds shows emergence at 10\% hierarchical-rule inducing data. \textit{Right-4}: Histograms reveal bimodal distributions at 10\%, 90\% and 100\% hierarchical data. Models either cluster at either 0\% (linear generalization) or 100\% (hierarchical generalization). 
    }
    \label{fig:qf_data}
\end{figure*}

\begin{figure}
    \centering
    \includegraphics[width=0.7\linewidth]{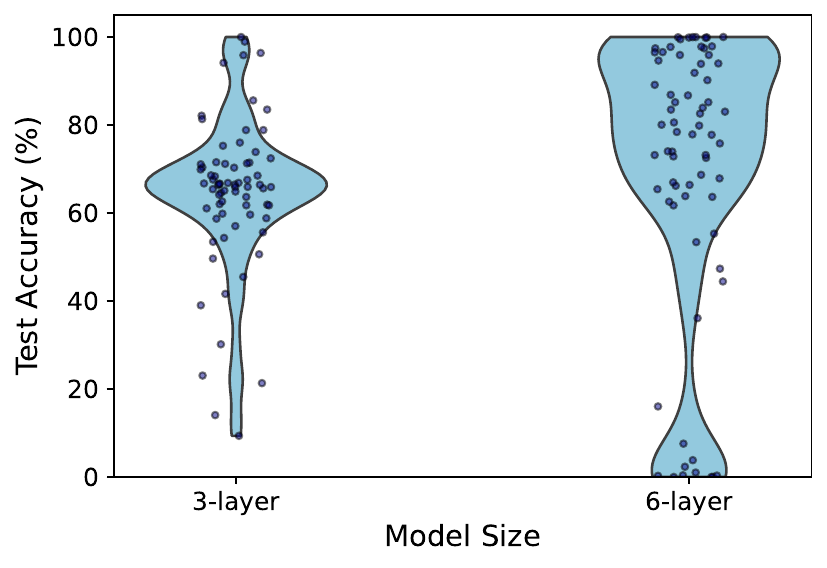}
    \caption{\textbf{Question formation performance across model scales.} Training on 90\% hierarchical-rule inducing data, 3-layer models show unimodal distribution around 70\% OOD accuracy, while 6-layer models exhibit bimodality with most seeds achieving 100\% (hierarchical generalization) and some reaching 0\% (linear generalization). Each dot represents one seed.}

    \label{fig:qf_scale}
\end{figure}

\begin{figure*}[t]
    \centering
    \begin{minipage}{0.48\textwidth}
        \centering
        \includegraphics[width=1.0\linewidth]{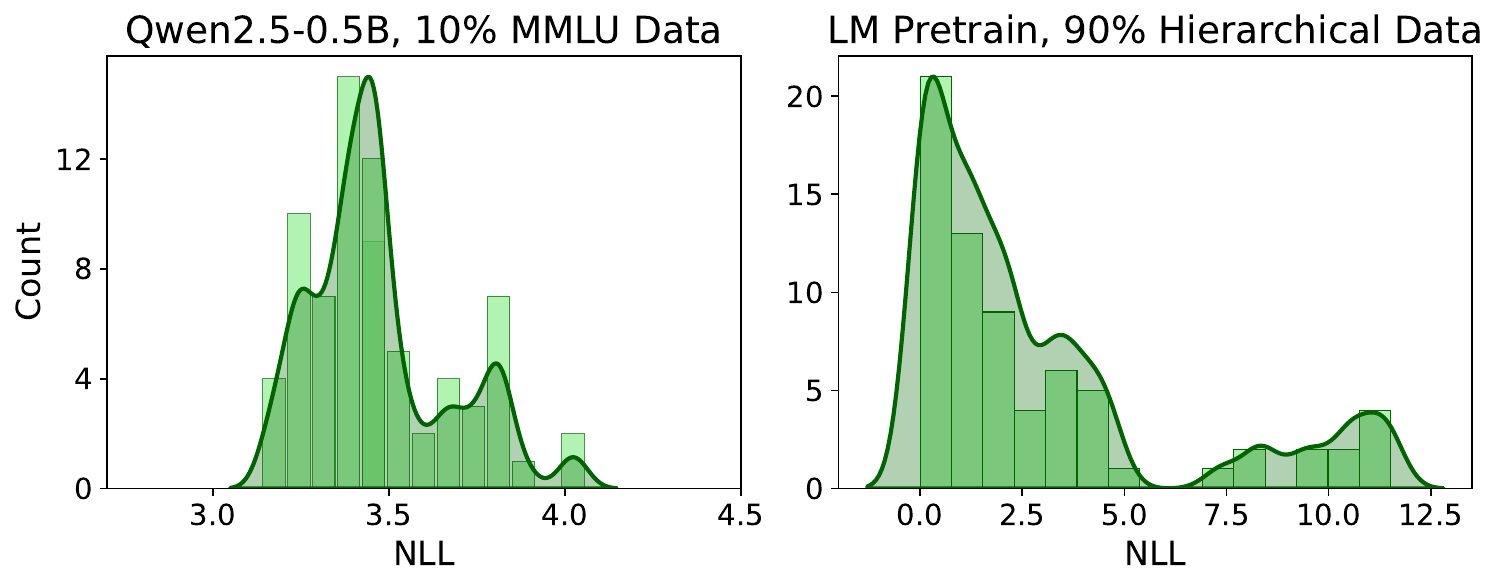}
        \caption{\textbf{NLL loss distributions.} Even with a \textbf{continuous} metric (NLL), we still observe multi-modal distributions in performance across random seeds. NLL is measured on: \textit{Left:} MMLU for Qwen2.5-0.5B trained on 10\% MMLU mixture. \textit{Right:} OOD QF for LM pretrained on 90\% hierarchical mixture.}
        \label{fig:qwen_mmlu_hist}
    \end{minipage}
    \hfill 
    \begin{minipage}{0.48\textwidth}
        \centering
        \vspace{-7px}
        \includegraphics[width=0.68\linewidth]{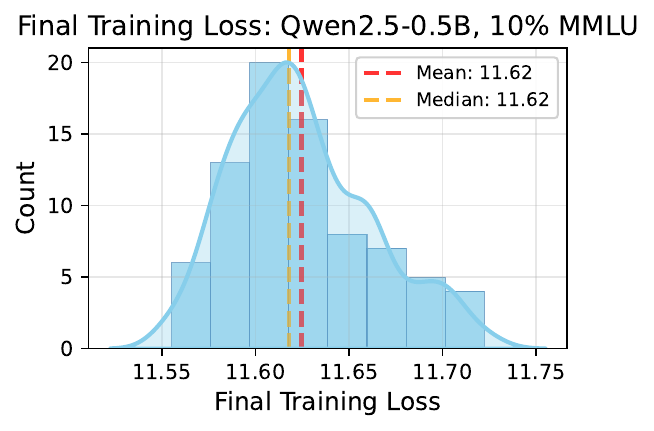}
        \vspace{-4px}
        \caption{\textbf{KDE of LM training loss.} The distribution across 80 seeds is unimodal and smooth. By contrast, Fig.~\ref{fig:qwen_mmlu_hist} depicts MMLU loss as irregularly clustered.}
        \label{fig:trainloss}
    \end{minipage}
    \vspace{-10px}
\end{figure*}

\subsection{Is MMLU emergence a mirage?}
\label{subsec:mmlu-mirage}

Is multiple choice performance only bimodal because accuracy is thresholded? As in our synthetic setting (Sec.~\ref{subsec:bimodality_mirage}), we examine continuous metrics. In Fig.~\ref{fig:qwen_mmlu_hist}, negative log likelihood (NLL) loss remains bimodal. For the 10\% MMLU training data case, we confirm this bimodality is statistically significant ($p < 0.001$) for  significant using Hartigan's Dip Test \citep{diptest}, which measures the maximum difference between the empirical distribution and the best-fitting unimodal distribution.
The bimodality observed in NLL also mirrors our synthetic algorithmic task findings. 

While recent work \citep{srivastava2023beyond,schaeffer2024emergent} argues that discontinuities disappear with continuous metrics, Fig.~\ref{fig:qwen_mmlu_hist_contrast} and Fig.~\ref{fig:qwen_mmlu_1.5b} reveals the exact opposite: NLL loss exposes clusters \textit{concealed} by accuracy metrics, particularly among low-performance models trained on 5\% MMLU (Fig.~\ref{fig:qwen_mmlu_hist_contrast}) and among larger models (Fig.~\ref{fig:qwen_mmlu_1.5b}).

Is this clustering specific to emergent tasks?  Fig.~\ref{fig:trainloss} suggests so.  Pretraining loss produces a smooth, unimodal, nearly symmetric distribution, contrasting sharply with MMLU's multimodality. We therefore conclude that multimodality is linked to emergent tasks. It is \textit{not merely} a downstream effect of in-distribution training noise.

\section{Grammatical Generalization in LMs}

\label{sec:grammar}

We next examine English syntax acquisition during LM training. Question formation (QF) is a canonical benchmark \citep{Mueller2023-xq, McCoyUnknown-uy, McCoy2020-pj} for grammatical rule learning. In this setting, the ambiguous training data supports two possible competing rules:  (i) a hierarchical syntactic rule (correct for English) or (ii) a superficial linear heuristic (like an n-gram language model). On the disambiguating OOD test data, the hierarchical rule yields ~100\% accuracy, the linear rule yields ~0\%, and any score in between indicates inconsistent rule application. 

Prior work has documented performance clustering in QF across random seeds \citep{qin2024itree, Ahuja2024-ul, Murty2023-xp}, with evidence that data composition and model scale influence which rule is learned \citep{qin2024itree, Ahuja2024-ul}. This controlled setting allows us to study bimodal distributions of competing viable solutions when training a simple language model from scratch.

\subsection{Data and Setup}

\textbf{Task}: We pretrain models on the \textbf{question formation} task \citep{McCoy2018-uv}, which tests English syntax acquisition. Models transform declarative sentences to questions by moving the main auxiliary verb (e.g., "My unicorn \textit{does} move" $\rightarrow$ "\textit{Does} my unicorn move?"). The training data contains some sentences that induce the hierarchical rule (through \textit{center embedding} syntactic structure) and some sentences that induce the linear rule (through \textit{right-branching} syntactic structure). Critically, the question formation examples in the training data are \textit{always ambiguous} between both rules. The OOD test data then probes for which rule has been learned. (Full task details in Appx.~\ref{appdx:qd_setup}.)

\textbf{Model}: Decoder-only Transformers trained from scratch on causal language modeling. Following prior work \citep{qin2024itree, Murty2023-xp}, we vary depth (3-6 layers) with 8 heads per layer and 512-dimensional embeddings. 

\textbf{Data}: Synthetic English sentences generated via Context-Free Grammars \citep{McCoy2018-uv}. Based on \citet{qin2024itree}, we vary the proportion of hierarchical-inducing versus linear-inducing sentences in the training data, paralleling our MMLU data mixture experiments.

\textbf{Training}: We train 80 random seeds per configuration. We use Adam optimizer with learning rate $1\times10^{-4}$ and linear decay and train models for 600K steps. We vary (1) model size and (2) proportion of hierarchy-inducing training examples, to study how these scale factors influence the bimodal performance distribution of hierarchical generalization.

\subsection{Experiment Results}
\label{sec:qf_result}
\paragraph{Emergence Across Data Compositions.}

Fig.~\ref{fig:qf_data} (\textit{left}) shows mean OOD accuracy exhibits sharp emergence at 10\% hierarchical data, jumping from near 0\% to 80\%. However, histograms (\textit{right 4 panels}) reveal that the population mean masks the underlying bimodality: at 10\%, seeds split between 0\% (linear rule) and 80-100\% (hierarchical rule). 
Varying data composition shifts \textit{which} rule models learn. More hierarchical-inducing sentences increase the probability of learning the hierarchical rule, but do not eliminate the competing solution even at 100\% hierarchical-inducing data. Apparently, even our most hierarchical-inducing training set offers insufficient support to provide consistently ``successful'' runs, possibly because it lacks diversity.

\vspace{-10px}
\paragraph{Emergence Across Model Scales.}

Fixing the data mixture at 90\% hierarchical sentences, we now examine how model capacity affects the performance distribution. At 3 layers (Fig.~\ref{fig:qf_scale}), performance is unimodal around 70\%, with few seeds reaching 100\%---these smaller models lack capacity to learn either systematic syntactic rule. At 6 layers, clear bimodality emerges: most seeds achieve 100\% (hierarchical generalization) while some reach 0\% (linear generalization). Although our largest scale models tend to consistently apply systematic rules, we find different seeds still vary in \textit{which} rule they apply. Finally, similar to Sec.~\ref{subsec:mmlu-mirage}, we confirm that the bi-modality is still observed under continuous metric in Figure~\ref{fig:qwen_mmlu_hist} and is statistically significant under Hartigan’s Dip Test \citep{diptest}.\looseness=-1

\section{Discussion}
Our work explores the evolution of random variation in model performance across scales, bringing a nuanced perspective on emergent capabilities. While the mode of ``emergent'' performance distributions may sharply improve at a certain model scale, we attribute these sudden jumps to gradual improvements in the random distribution. In fact, bimodality often emerges \emph{before} the mode---or most individual runs---exhibits a breakthrough, even when the transition from a unimodal distribution to a bimodal one is sudden.

\paragraph{Random variation:} Model performance can be sensitive to stochastic aspects of the training process like random initialization and training data order. While variation may have a benign effect on in-distribution performance~\citep{jordanvariance}, previous works indicate that out-of-distribution performance can fluctuate substantially across training runs, even at larger scales~\citep{madaan2024quantifying}.
Prior studies have documented  performance differences across various stress test sets~\citep{d2022underspecification, naik2018stress}, including length generalization~\citep{zhou2024transformers, zhoualgorithms}. More generally, out-of-distribution behavior like compositional rules \citep{mccoy_berts_2019} or associative biases  \citep{sellammultiberts} often exhibit extreme variation compared to in-distribution loss. Such differences persist throughout training, not just at the final checkpoint~\citep{zhou2020curse}. \citet{dodge2020fine} compared the impacts of weight initialization and data ordering, concluding that both contribute equally to variation in performance. 

Existing work has also found model runs can cluster in shortcut learning \citep{juneja_linear_2023,li2025interpretationpredictbehaviorunseen} and in training dynamics \citep{qin2024itree,hu_latent_2023}, hinting at multimodal variation. Our work connects these random clustering effects to the phenomenon of emergence at scale.  Relevant to our setting, \citet{zhoualgorithms} and \citet{zhou2024transformers} provided evidence of variability in length generalization across random seeds, which we further analyze across model scales. We also expand on the findings of \citet{qin2024itree} of bimodal variation in LM syntax acquisition.


\paragraph{Emergent abilities of LMs:} In LMs, scaling laws predict reliable performance improvements as  models increase in size or train on larger datasets~\citep{hestness2017deeplearningscalingpredictable,rosenfeld2019constructivepredictiongeneralizationerror,brown2020language,kaplan2020scaling}.  Emergent abilities are abilities that arise unexpectedly, out of line with these predictions \citep{srivastava2023beyond,ganguli2022predictability,wei2022emergent}. These abilities are characterized by unpredictable and abrupt performance improvements on specific benchmarks at certain scales. Although recent studies suggest that some breakthroughs may stem from the choice of evaluation metrics rather than fundamental changes in model behavior~\citep{srivastava2023beyond,schaeffer2024emergent}, other breakthrough capabilities remain emergent---8\% of high-breakthroughness datasets in BIG-Bench, according to \citet{schaeffer2024emergent}. Our findings suggest that careful statistical analysis is required to evaluate claims both for and against ``emergence'' scaling on specific tasks.

\citet{snell2024predicting} found that some scales exhibit earlier emergence if finetuned explicitly on an emergent task, suggesting that smaller models may have the capacity for that task but are limited by its scarcity in the training corpus. Similarly, we show that emergent capabilities can arise from multimodal random variation using synthetic length generalization tasks as a case study.

\paragraph{Depth versus Width Scaling:} Downstream performance varies with architecture shape, not just model size~\citep{tayscale}. For compositional tasks, deeper models often generalize better, but for fixed compute budgets, shallower and wider models may be advantageous~\citep{petty2024impact}. Several explanations of this benefit have been proposed. \citet{edelman2024pareto} argued that wider networks offer more parallel queries over randomized subnetworks and therefore learn sparse features efficiently. \citet{levine2020limits} established width-dependent depth thresholds beyond which depth yields diminishing returns~\citep{levine2020limits}. We investigate how independently scaling width and depth influences random variation in compositional tasks, documenting a surprising regime (Appx. \ref{subsec:inverse_scaling}) where increasing width damages performance while increasing depth improves it. Our findings should inspire further study of how emergent tasks respond to architectural hyperparameter tradeoffs.

\section*{Acknowledgments}

This work has been made possible in part by a gift from the Chan Zuckerberg Initiative Foundation to establish the Kempner Institute for the Study of Natural and Artificial Intelligence. RZ, TQ, and SK acknowledge support from the Office of Naval Research under award N00014-22-1-2377 and the National Science Foundation Grant under award \#IIS 2229881. RZ is supported by a Kempner Institute Graduate Research Fellowship, Simons Investigator Fellowship, NSF grant DMS-2134157, DARPA grant W911NF2010021,and DOE grant DE-SC0022199. TQ and DAM are partially supported by the Kempner Institute, the Aramont Fellowship Fund, and the FAS Dean’s Competitive Fund for Promising Scholarship. Our work has been improved by invaluable discussion with Will Merrill and David Chiang.

\bibliography{references}
\bibliographystyle{icml2026}

\newpage
\appendix

\clearpage
\onecolumn
\section{Experimental details on synthetic tasks}
\label{app:exp_details}

Below we provide more details on the \textbf{count} and \textbf{addition} task settings. Hyperparameters for our decoder-only Transformer models are largely adapted from~\citep{zhoualgorithms}. We train all of our models to convergence on the train distribution. Our experiments were run on an internal cluster and all model scales can be run on a single 40GB A100 GPU with gradient accumulation; for \textbf{count}, runs can finish within 2 hours and for \textbf{addition}, runs can finish within 6 hours. At each  scale, we train models from 250 seeds for the count task and 200 seeds for reverse order addition.  

\textbf{Count task: } For all of our training runs, we fix the vocabulary size to 150. For thresholded evaluation, we compute the exact match (EM) accuracy across all consecutive subsequences of the test length. 

\begin{itemize}
\item \textbf{Model scales:} As mentioned in Sec.~\ref{sec:methodology_synthetic}, we scale up our models by fixing width and scaling depth and fixing depth and scaling width. The precise parameters for each variation are as follows. For our \textbf{fixed depth experiments}, we fix the network depth to 4 layers and vary width by taking hidden dimensions $\{64, 128, 256, 384, 512, 640, 768, 1024\}$. The head dimension is fixed to 64. For our \textbf{fixed width experiments}, we fix the hidden dimension to be 512 and vary the depth from $\{1, 2, 4, 6, 8\}$ layers.
\item \textbf{Hyperparameters:}  We use a learning rate of $1e-3$ with a cosine decay scheduler and weight decay 0.1. We set the maximum training duration to be 10000 steps, with batch size 128 and context length 256. 
\end{itemize}

\textbf{Reverse Order Addition with Index Hints:} For thresholded evaluation, we compute the exact match (EM) accuracy across 500 batches of 128 examples each.

\begin{itemize}
\item \textbf{Model scales:} For our \textbf{fixed depth experiments}, we fix the network depth to 6 layers and vary width by taking hidden dimensions $\{64, 128, 256, 384, 512, 640, 768\}$. For our \textbf{fixed width experiments}, we fix the hidden dimension to be 512 and vary the depth from $\{1, 2, 3, 4, 6, 8, 10, 12\}$ layers.
\item \textbf{Hyperparameters:}  We use a learning rate of $1e-4$ with a cosine decay scheduler and weight decay 0. We set the maximum training duration to be 30000 steps, with batch size 64 and context length 512. 
\end{itemize}

\section{Breakthroughness and Linearity}
\label{app:breakthrough}

\citet{srivastava2023beyond} introduced \emph{breakthroughness} and \emph{linearity} metrics to capture model performance improving suddenly or reliably with scale. Given a model's performances $y_i$ at model scales $x_i$ sorted by ascending model scale, the linearity metric $L$ and breakthroughness metric $B$ are respectively calculated as
\begin{align}
    &L = \frac{I(y)}{\mathrm{RootMeanSquare}(\{y_{i + 1} - y_i\}_i)} \\
    &B = \frac{I(y)}{\mathrm{RootMedianSquare}(\{y_{i + 1} - y_i\}_i)}
\end{align}
where $I(y) = \mathrm{sign}(\arg\max_i y_i - \arg\min_i y_i) (\max_i y_i - \min_i y_i)$.

\begin{figure*}
    \centering
    \begin{subfigure}{0.7\textwidth}
\centering
\includegraphics[width=\linewidth]{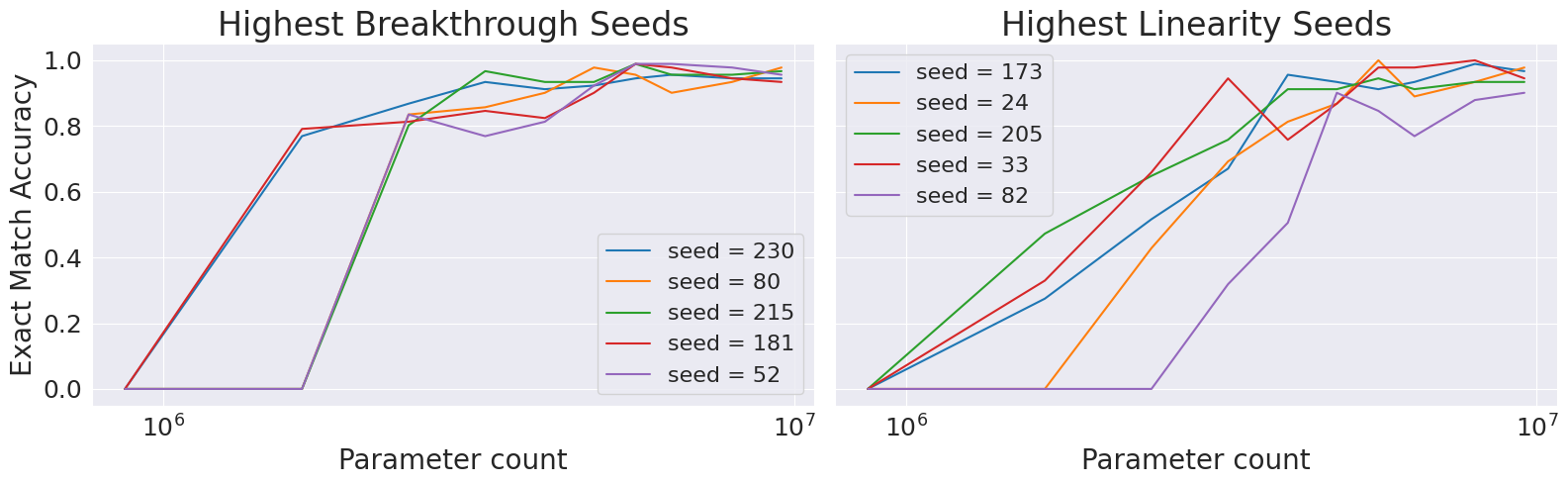}
\caption{Count.}
\label{fig:count_breakthrough}
\end{subfigure}
    \begin{subfigure}{0.7\textwidth}
\centering
\includegraphics[width=\linewidth]{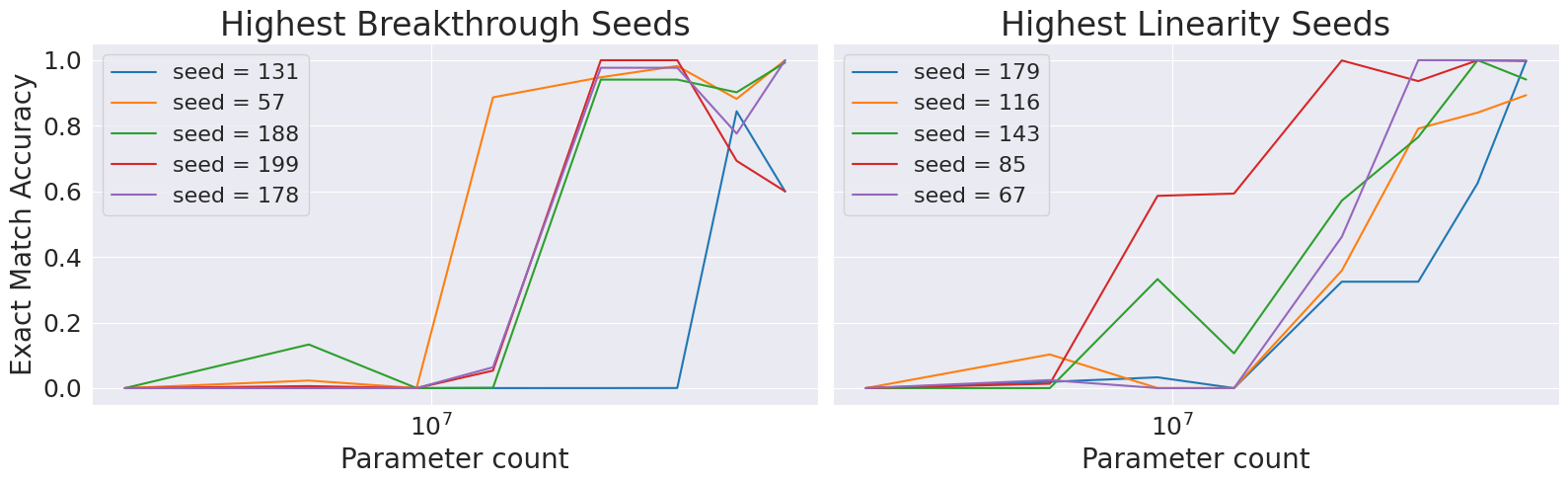}
\caption{Addition.}
\label{fig:rev_add_breakthrough}
\end{subfigure}
    \caption{\textbf{Top five seeds according to breakthroughness and linearity metrics.} Definitions for the two metrics are given in Appx.~\ref{app:breakthrough}, with resulting seeds plotted for \protect{\subref{fig:count_breakthrough}} count  and \protect{\subref{fig:rev_add_breakthrough}} addition.}
    \label{fig:breakthrough_seeds}
\end{figure*}

In Fig.~\ref{fig:breakthrough_seeds} we sample the five top seeds for the breakthroughness and linearity metric respectively for (a) count and (b) addition.

\section{Count task}
\label{app:count}

Given two numbers in increasing order, the model is trained to generate a sequence which counts consecutively from the first number to the second number. Examples are given in the form \texttt{"5, 9 >, 5, 6, 7, 8, 9"}, while limiting the length of the counting sequence during training. 

\citet{zhoualgorithms} showed that models trained to count can generalize to more than twice this training length; however, Appx. \ref{subsec:inverse_scaling} reveals a more nuanced view of length generalization based on its distribution across independent model runs.

\subsection{Count performance distributions}
\label{app:add_em_histograms}

\begin{figure*}[ht]
    \centering
    \begin{subfigure}{\linewidth}
\centering
\includegraphics[width=\linewidth]{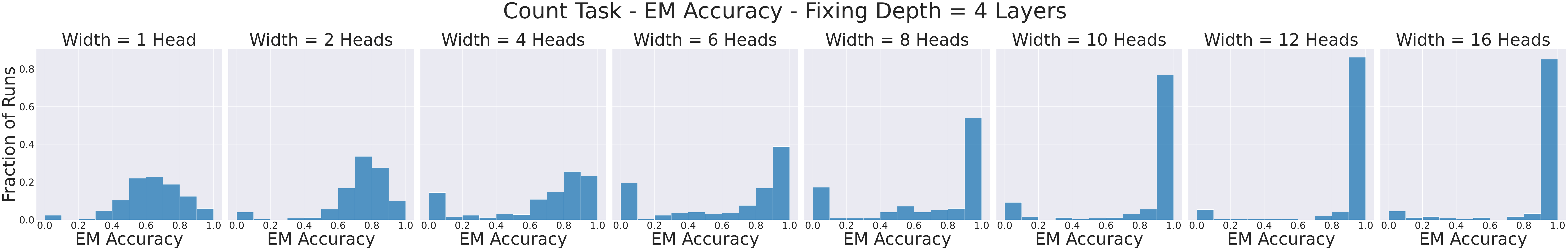} 
\caption{Fixing network depth to 4 layers.}
\label{fig:count_fix_depth_hist}
\end{subfigure}
    
    \vspace{0.2cm}  
    
\begin{subfigure}{\linewidth}
\centering
\includegraphics[width=\linewidth]{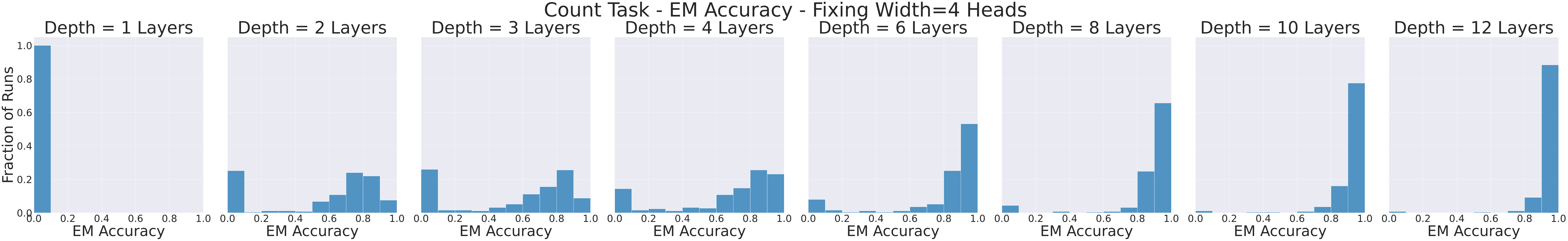} 
\caption{Fixing network width to 4 heads per layer.}
\label{fig:count_fix_width_hist}
\end{subfigure}
    \caption{\textbf{Random variation in length generalization (count task).} Histograms of exact match accuracy on length $60$ sequences when independently scaling (\protect{\subref{fig:count_fix_depth_hist}}). width  and (\protect{\subref{fig:count_fix_width_hist}}). depth.
    }
    \label{fig:count_histograms}
\end{figure*}

\begin{figure*}
    \centering
    \includegraphics[width=\textwidth]{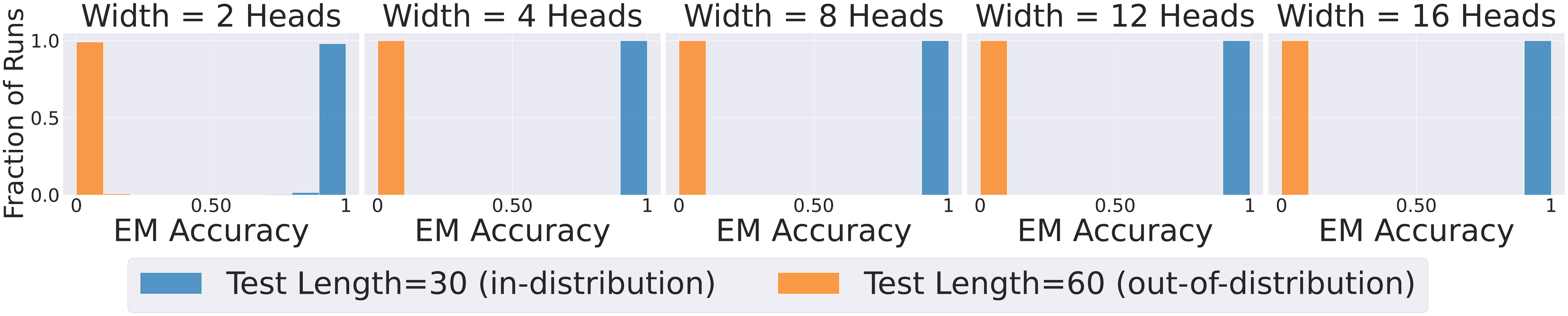}
    \caption{
    \textbf{Histograms for EM accuracy when fixing depth to be one layer (count task).} We plot the random variation distribution at test length 30 (blue, i.e. in-distribution) and test length 60 (orange, i.e. out-of-distribution). While all model seeds obtain near perfect accuracy in-distribution, all model seeds fail to length generalize at this depth.
    }
    \label{fig:count_fix_depth_1}
\end{figure*}

Our plots in the main paper show distributions of addition task accuracy with a fixed test length 40 (Fig.~\ref{fig:rev_add_histograms}). In Fig.~\ref{fig:count_histograms} we show analogous histograms for the count task with a fixed test length of 60. We note that unlike addition, the mode corresponding to failing runs disappears at the highest widths and depths ran, indicating a difference in task difficulty affecting the random variation distribution.

In Sec.~\ref{subsec:emergence_of_bimodality}, we posit that the performance distribution becomes bimodal at the \textbf{minimum capacity} required by the task. For example, Fig.~\ref{fig:count_fix_depth_1} shows that, regardless of width, models with a fixed depth of 1 layer are unable to length generalize on the count task, despite achieving near-perfect accuracy in-distribution.

\subsection{Count task as a case  of U-shaped random scaling}
\label{subsec:inverse_scaling}

\begin{figure}[ht!]
    \centering
    \includegraphics[width=0.6\linewidth]{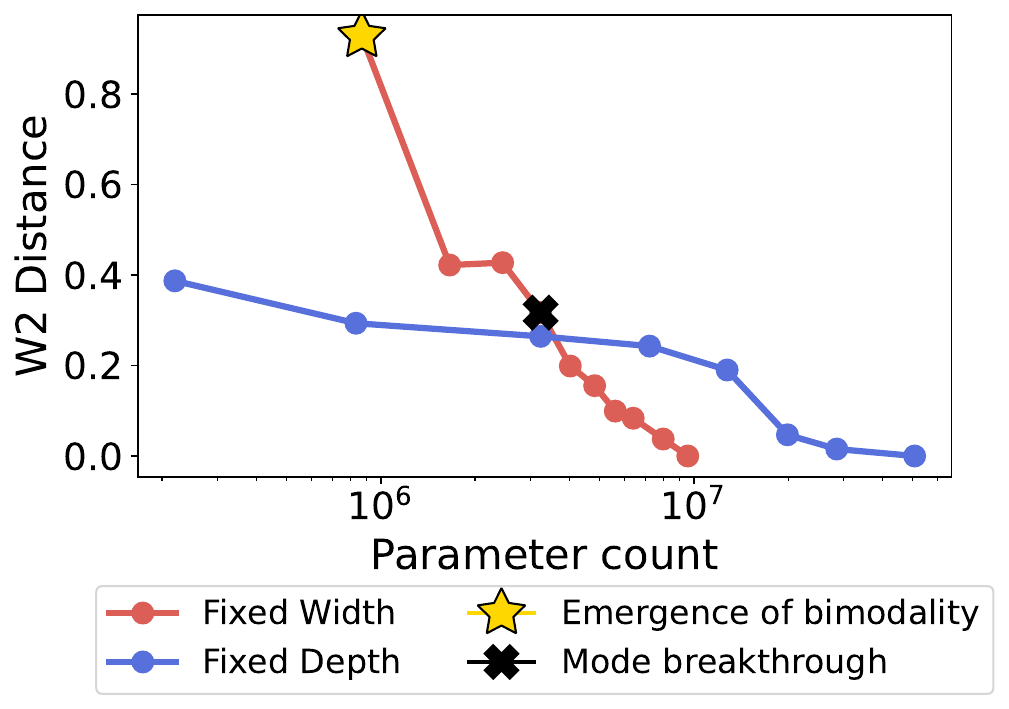}
\caption{\textbf{Changes in random variation (count task).} Wasserstein-L2 distance of each scale's performance distribution relative to the largest scale, scaling depth and width independently.  We mark  the emergence of bimodality at the last scale before multiple peaks appear. We mark the mode breakthrough  at the last scale before successful length generalization becomes marginally more likely than failure. \textit{The distribution changes slowly between intermediate model scales, but changes suddenly at the nadir of the U-shaped curve in Fig. \ref{fig:count_quantiles}.}}
    \label{fig:count_wasserstein}
\end{figure}

\begin{figure}[t!]
    \begin{subfigure}[t]{\linewidth}
        \centering
        \includegraphics[width=0.8\linewidth]{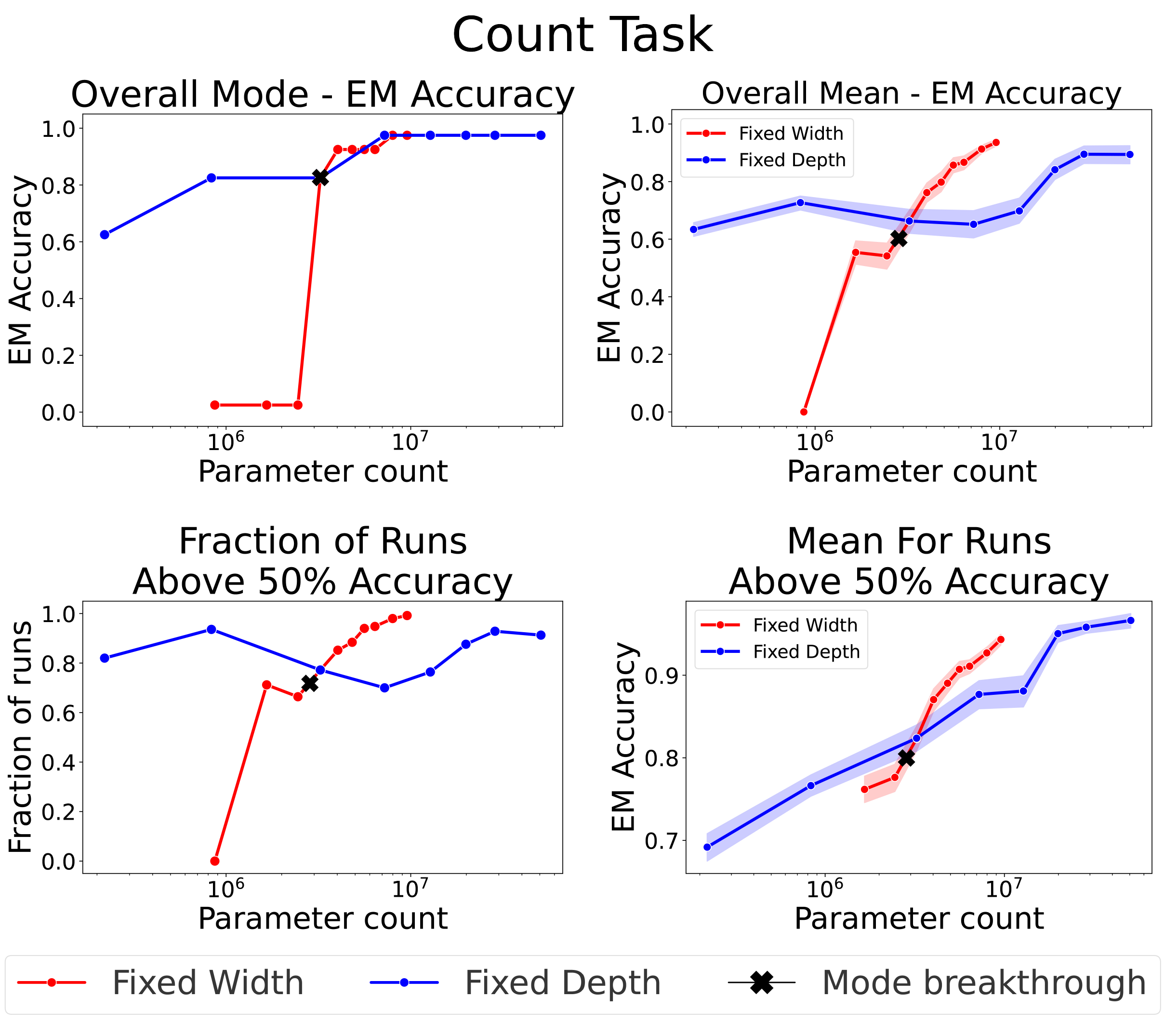}
    \end{subfigure}
    \caption{\label{fig:count_quantiles} As in Fig.~\ref{fig:rev_add_quantiles}, we capture summary statistics for test length 60 on the count task. For fixed width (scaling depth), the trends across overall mode, overall mean, fraction, and mean of successful (above 50\% accuracy) runs are similar to the reverse order addition case. The trend for fixed depth (scaling width) exhibits a U-shaped curve emerge for the mean across \textit{all} runs; however, the mean of  \textit{successful} runs still exhibits continuous improvement.
    }
    \vspace{-0.1cm}
\end{figure}

We next consider the counting task (Sec. \ref{sec:methodology_synthetic}), which also exhibits bimodally-distributed performance (see Appx. Fig.~\ref{fig:count_histograms}) but  yields a very different scaling effect: a U-shaped curve.
 Fig.~\ref{fig:count_quantiles} reveals this peculiar phenomenon in the mean accuracy scaling curve, when holding depth fixed. This curve is not simply a result of the summary statistic chosen, as it is mirrored by the evolution of the distributions as a whole according to their W2 distance in Fig.~\ref{fig:count_wasserstein}. 

U-shaped scaling has been observed in LMs, but its causes are not currently well-understood. When the Inverse Scaling Prize~\citep{mckenzie2022inverse} solicited tasks which exhibit inverse scaling trends---performance decreasing with scale---for large models,  \citet{wei2023inverse} revealed that the majority of awarded tasks actually exhibit U-shaped scaling after considering even larger models.
Treating the counting task as a concrete instance of U-shaped scaling at small model scales, we find that this unusual trend is still underscored by monotonic continuous changes in the performance distribution. Indeed, Fig.~\ref{fig:count_quantiles} (bottom right) shows that although the trend in the mean across all runs is U-shaped curve, the mean of the ``successful'' runs---those achieving at least 50\% accuracy---still improves monotonically when increasing width. The observation of inverse scaling is, instead, due to changes in the \textit{probability} of success (bottom left). Even when inverse scaling is in effect across a performance distribution, the performance of  successful runs may exhibit more conventional responses to scale.

\section{Additional Continuous Metric Plots}
\label{app:add_continuous_metrics}

\paragraph{Addition Task.} In Sec.~\ref{subsec:bimodality_mirage} we have seen seen that the addition histograms are bimodal in exact match accuracy (Fig.~\ref{fig:rev_add_histograms}) and confirmed that the bimodal random variation distribution persists when viewing a continuous performance metric (Fig.~\ref{fig:combined_results}~\textit{a, b} ). We can consider another continuous metric, which corresponds to the minimum probability assigned to a token for each sample in the test length generalization dataset:

\begin{align}
\label{eq:probability}
    \textrm{minprob}(f, X) = \frac{1}{|X|} \sum_{x \in X} \min_{i < |x|} \mathbf{P}_{f}[x_i | x_{0 \ldots i-1}]
\end{align}

\begin{figure*}
    \centering
    \begin{subfigure}{\linewidth}
\centering
\includegraphics[width=\linewidth]{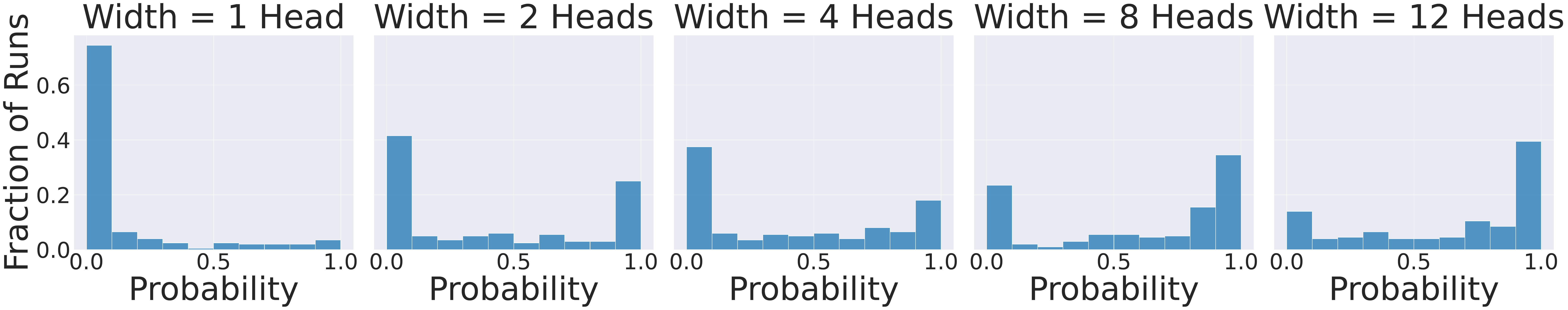}
\caption{Fixing network depth to 6 layers while scaling up width.}
\label{fig:rev_add_histograms_probability:depth}
\end{subfigure}
    \begin{subfigure}{0.6\linewidth}
\centering
\includegraphics[width=\linewidth]{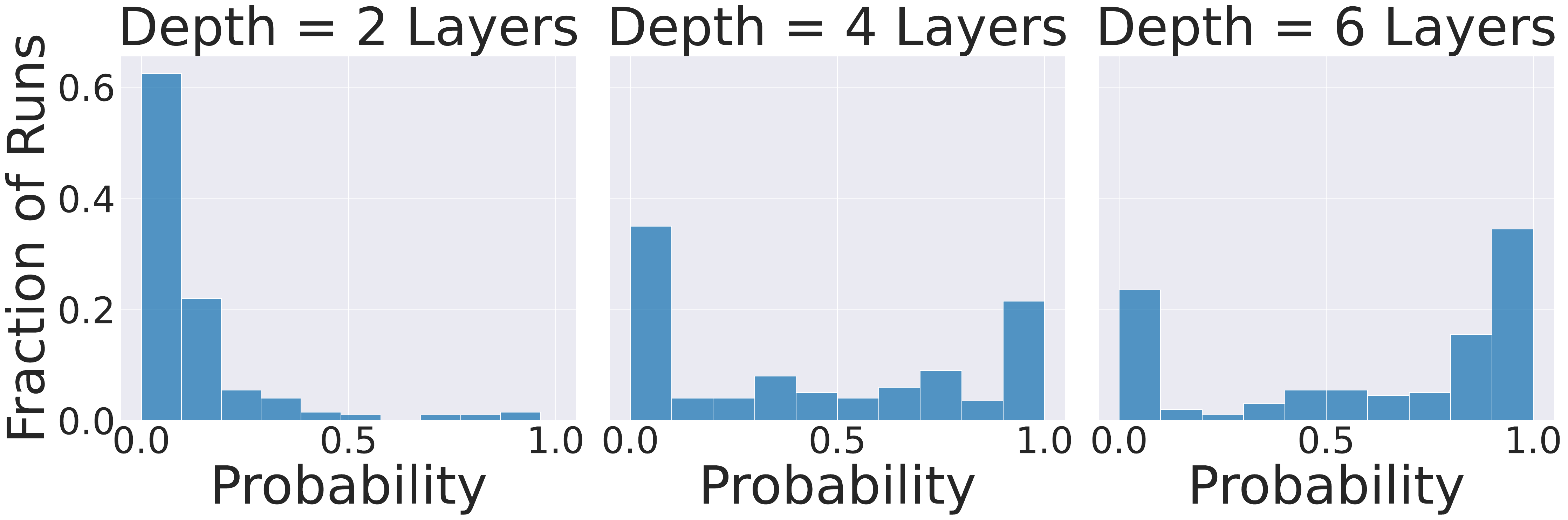}
\caption{Fixing network width to 8 heads per layer while scaling up depth.}
\label{fig:rev_add_histograms_probability:width}
\end{subfigure}
    \caption{\textbf{Histograms of the minimum probability of any token in each sequence, averaged across sequences (addition task).} Metrics are calculated by Equation \ref{eq:probability}. The bimodal nature of the random variation distribution persists even when using this continuous metric, analogous to Fig.~\ref{fig:combined_results}~(a, b).}
    \label{fig:rev_add_histograms_probability}

\end{figure*}

\begin{figure*}
    \centering

    \begin{subfigure}{\linewidth}
\centering
\includegraphics[width=\linewidth]{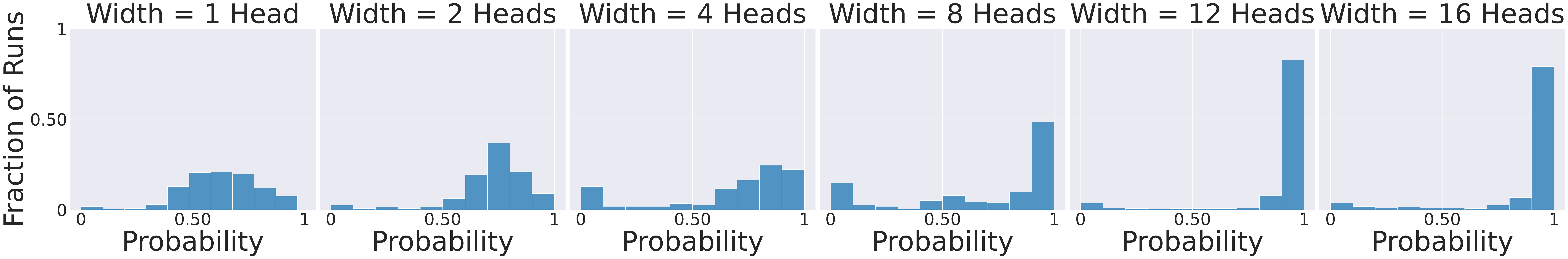}
\caption{Fixing network depth to 4 layers.}
\end{subfigure}
    \begin{subfigure}{0.7\textwidth}
\centering
\includegraphics[width=\linewidth]{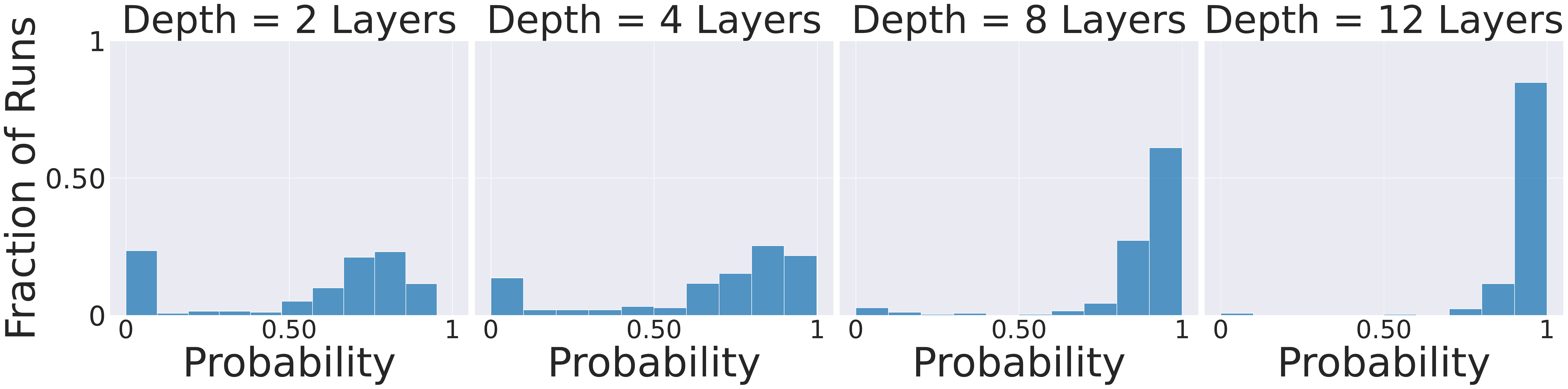}
\caption{Fixing network width to 256 hidden dimension.}
\end{subfigure}
        
    \caption{\textbf{Histograms of the minimum probability of any token in each sequence, averaged across sequences (count task).} Random variation still leads to bimodal performance distributions, even  using this continuous performance metric.}
    \label{fig:count_histograms_probability}

\end{figure*}

We consider the average minimum probability (and analogously average maximum loss) in the sequence because errors are rare, and thus averaging across the sequence obscures generalization failures. However, as the model improves, the lowest-probability token may shift, but this transition is still continuous, not abrupt. Thus, unlike 0/1 accuracy, this metric also avoids threshold effects and better reflects gradual improvements in length generalization, similar to the continuous error metric. We plot the resulting histograms of this metric in Fig.s~\ref{fig:rev_add_histograms_probability} and \ref{fig:count_histograms_probability} for addition and count respectively. We note that bimodality persists for both tasks, particularly for addition; in particular, stronger bimodality exhibited in the accuracy histograms corresponds to stronger bimodality in the probability histograms.

\begin{figure*}
    \centering
    \begin{subfigure}{\linewidth}
\centering
\includegraphics[width=\linewidth]{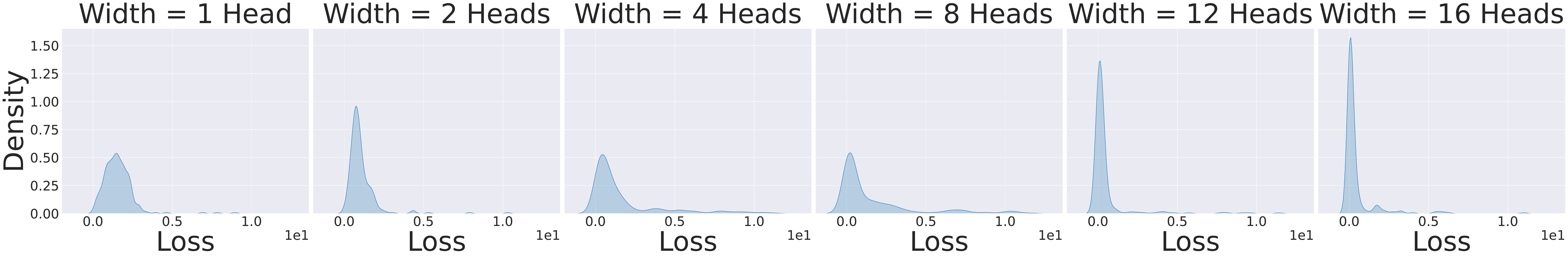}
\caption{Fixing network depth to 4 layers while scaling up width.}
\label{fig:count_histograms_loss:depth}
\end{subfigure}
    \begin{subfigure}{0.6\linewidth}
\centering
\includegraphics[width=\linewidth]{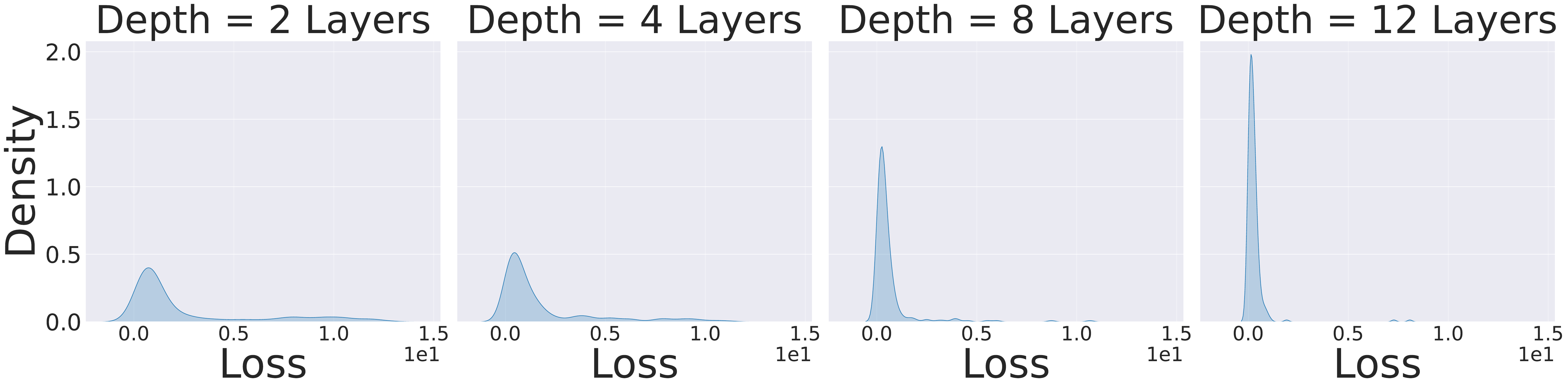}
\caption{Fixing network width to 6 heads per layer while scaling up depth.}
\label{fig:count_histograms_loss:width}
\end{subfigure}
    \caption{\textbf{Random variation in continuous length generalization error (count task).} Kernel Density Estimation (KDE) of loss-based error metric (Equation \ref{eq:continuous_error}) distribution across model runs. At scales where the EM accuracy distribution is most strongly bimodal (width=4, width=8, and depth=2),  KDE places the most density at areas of high loss.}
    \label{fig:count_histograms_loss}

\end{figure*}

\paragraph{Count Task.} In Fig.~\ref{fig:count_histograms_loss}, we plot the analogous KDE plot of the loss-based error metric for the count task. For count, we note that the mode corresponding to `failing runs' when looking at accuracy or probability is more diffuse compared to addition. However, we claim that stronger bimodality in the original accuracy/probability plots in Fig.s~\ref{fig:count_histograms} and \ref{fig:count_histograms_probability} is still associated with stronger bimodality in the loss plots. For the addition task, the scales exhibiting the strongest bimodality in accuracy and probability are for width 4 and 8  (in the case of fixed depth) and depth 2 and 4 (in the case of fixed width), and these are also the most strongly bimodal loss KDE distributions. For the count task, even the most strongly bimodal setting (fixed width, with depth 2) has a low probability of failure (~25\%), but still produces a visible wide peak elevated over the long tail of the loss distribution.

\begin{figure*}[h]
    \centering
    \includegraphics[width=0.8\linewidth]{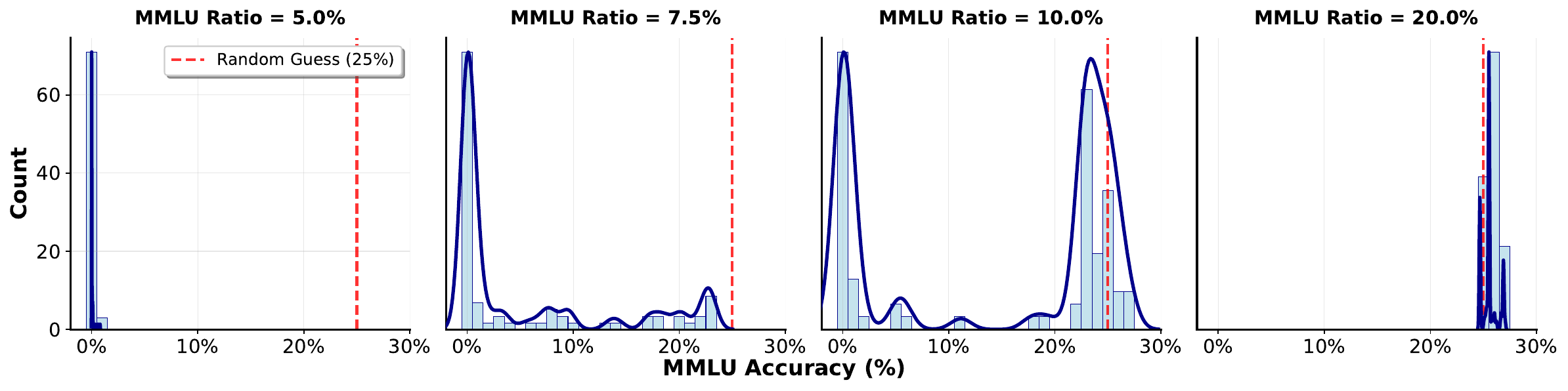}
     \includegraphics[width=0.8\linewidth]{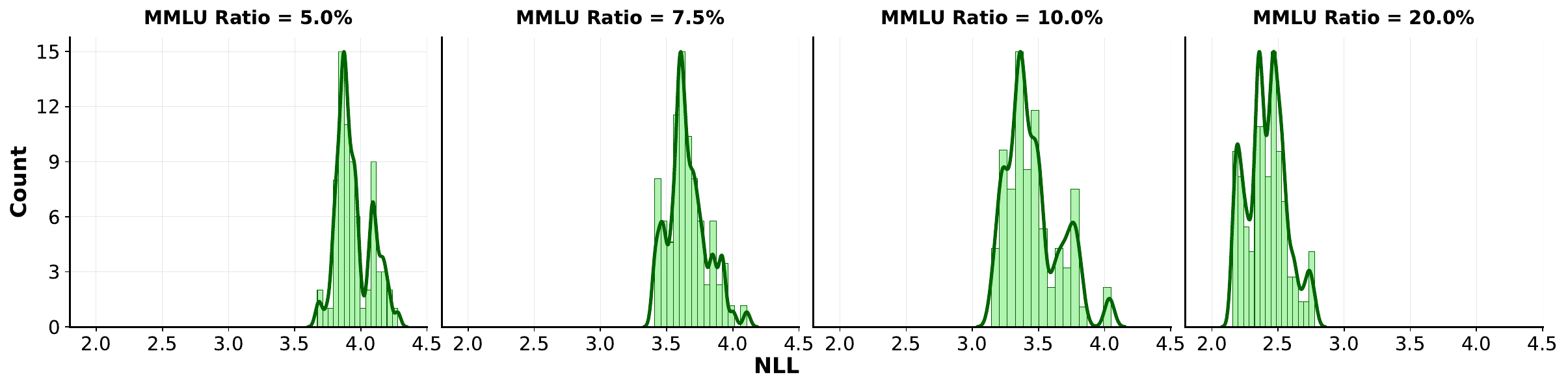}
    \caption{\textbf{KDE approximations of performance variation on the MMLU test set.} Provided for contrasting accuracy and continuous NLL loss. \textit{Top:} Approximate distribution of accuracy metrics. \textit{Bottom: } Approximate distribution of KDE metrics, copied from Fig. \ref{fig:qwen_mmlu_hist}.}
    \label{fig:qwen_mmlu_hist_contrast}
    \centering
    \includegraphics[trim=0cm 2cm 0cm 0cm, width=0.8\linewidth]{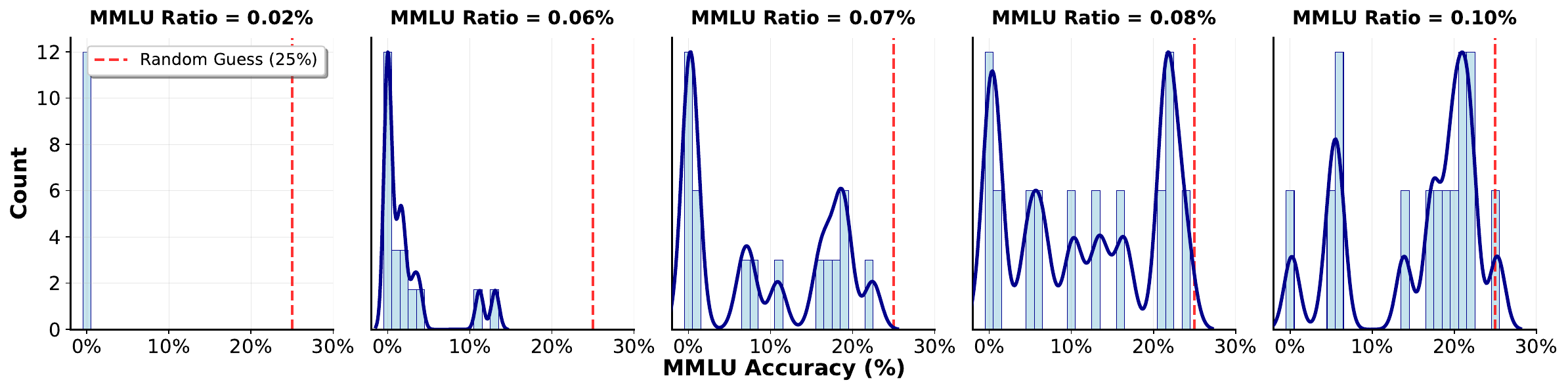}
     \includegraphics[width=0.8\linewidth]{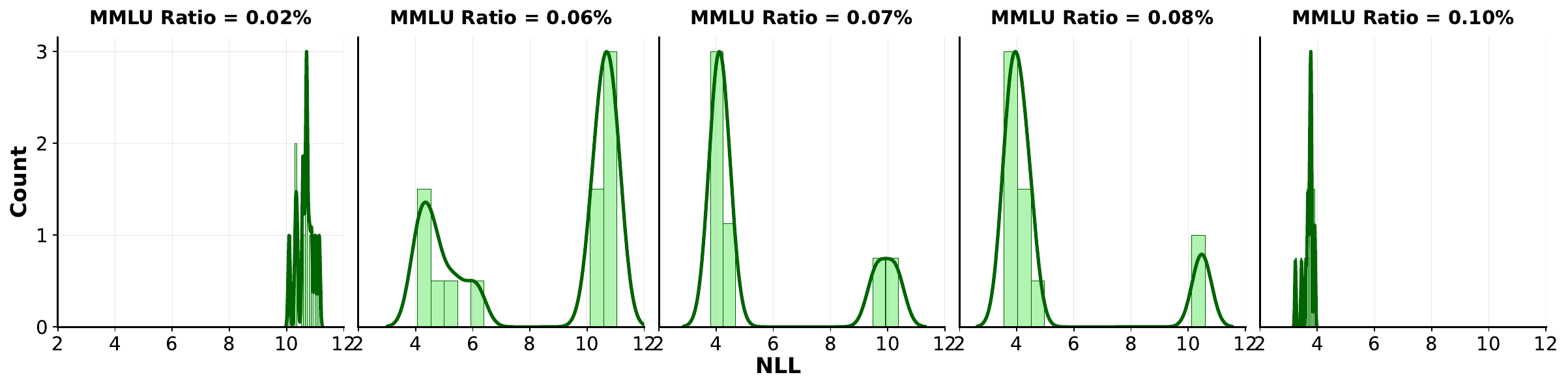}
    \caption{\textbf{Accuracy and NLL loss for Qwen2.5-1.5B MMLU experiments.} We train last-layer reinitialized Qwen2.5-1.5B on data mixes with different ratios of the target task (MMLU). \textit{Top:} Distribution of accuracy metric. \textit{Bottom: }Distribution of NLL loss metric.}
    \label{fig:qwen_mmlu_1.5b}
\end{figure*}

\paragraph{Large LM Experiments.} Finally, Fig. \ref{fig:qwen_mmlu_hist_contrast} provides a copy of Fig. \ref{fig:qwen_mmlu_hist} with both accuracy and NLL loss comparable for easy contrast. We note that high-performance clusters are distinguishable in both accuracy and loss, but the thresholded accuracy conceals low-performance clusters which are visible in the continuous distribution. We also include histogram version of results in Fig. \ref{fig:qwen_mmlu_1.5b}. Due to computation limit, for each data mixes, we train on 20 random seeds. Similar to Fig. \ref{fig:qwen_mmlu_hist_contrast}. models yields the same bimodal distribution in accuracy metrics as well as multi-modal distribution in continuous metric (NLL).

\section{Grammatical Generalization Details}
\label{appdx:qd_setup}
\subsection{Task Description}

In the \textbf{question formation (QF)} task, models transform declarative English sentences into questions by moving an auxiliary verb to the front of the sentence. The task tests whether models learn the correct hierarchical syntactic rule based on the sentence's syntax tree structure, or instead rely on a superficial linear heuristic.

Our training data (based on \citet{McCoy2018-uv}) permits two strategies for choosing which verb to move:

\begin{enumerate}[noitemsep,topsep=0pt,parsep=0pt,partopsep=0pt]
    \item \textbf{Linear rule}: Move the first auxiliary verb in the sentence (incorrect for English)
    \item \textbf{Hierarchical rule}: Move the main auxiliary verb, determined by the sentence's syntactic structure (correct for English)
\end{enumerate}

Examples of each rule are provided in Table \ref{tab:task_examples}. The first example is \textbf{ambiguous} because both the hierarchical and linear rules produce the same correct output. In contrast, the second example is \textbf{unambiguous} because only the hierarchical rule produces the grammatically correct question.

\subsection{Data Composition}

The training data contains two components:

\begin{enumerate}[noitemsep,topsep=0pt,parsep=0pt,partopsep=0pt]
    \item \textbf{Question formation examples}: Transform declarative sentences to questions. These examples are always ambiguous between linear and hierarchical rules.
    \item \textbf{Declaration copying examples}: Simply repeat declarative sentences without transformation. Unlike QF examples, these can include unambiguous sentence structures.
\end{enumerate}

Following \citet{qin2024itree}, we manipulate the declaration copying subset to control exposure to different syntactic structures. Specifically, we vary the ratio of \textbf{center-embedded sentences} versus \textbf{right-branching sentences}:

\begin{itemize}[noitemsep,topsep=0pt,parsep=0pt,partopsep=0pt]
    \item \textbf{Center-embedded sentences} contain recursive clause embedding (e.g., ``My unicorn who doesn't sing does move''), which requires hierarchical syntactic processing and encourages learning the hierarchical rule.
    \item \textbf{Right-branching sentences} have simpler linear structure (e.g., ``My unicorn does move the dogs that do wait''), which can be solved with linear n-gram patterns.
\end{itemize}

We create training sets with different mixtures of these sentence types in the declaration copying subset: \textit{Quest Only} (no declaration copying), \textit{Center Embed} (only center-embedded declarations), \textit{Right Branch} (only right-branching declarations), and various intermediate mixtures.

\subsection{Evaluation}

The in-distribution test set contains only ambiguous examples (solvable by both rules), while the OOD test set contains only unambiguous examples (requiring the hierarchical rule). Models using the hierarchical rule achieve $\sim$100\% accuracy on both test sets. Models using the linear rule achieve $\sim$100\% on in-distribution data but $\sim$0\% on OOD data. We therefore use \textbf{OOD accuracy} as our measure of whether models have acquired the hierarchical syntactic rule.

\begin{table*}[h]
    \centering
    \small
    \caption{\textbf{Question Formation task examples.} Models must move the main auxiliary verb to the front to form a question. Ambiguous examples can be solved by either linear or hierarchical rules; unambiguous examples require the hierarchical rule.}  
    \label{tab:task_examples}
    \begin{tabular}{p{2.5cm} | p{2.3cm} | p{9cm}}
    \hline
    \textbf{Task} & \textbf{Example Type} & \textbf{Examples} \\ 
    \hline
    \multirow{2}{*}{\shortstack{Question\\Formation}} & Ambiguous  & \textbf{Input:} My unicorn \textcolor{ForestGreen}{does} move the dogs that \textcolor{red}{do} wait. \\ 
    &  & \textbf{Output:} \textcolor{ForestGreen}{Does} my unicorn move the dogs that \textcolor{red}{do} wait? \\
    \cline{2-3} 
    & \multirow{3}{*}{Unambiguous} & \textbf{Input:} My unicorn who \textcolor{red}{doesn't} sing \textcolor{ForestGreen}{does} move. \\ 
    & & \textbf{Linear Output:} \textcolor{red}{Doesn't} my unicorn who sing \textcolor{ForestGreen}{does} move? \\
    & & \textbf{Hierarchical Output:} \textcolor{ForestGreen}{Does} my unicorn who \textcolor{red}{doesn't} sing move? \\
    \hline
    \end{tabular}
\end{table*}

\end{document}